\documentclass{article}
\usepackage{arXiv/arxiv}
\AtBeginDocument{
  \setlength{\headheight}{29pt}
  \addtolength{\topmargin}{-15pt}
}

\usepackage[T1]{fontenc}
\usepackage{mathpazo}               
\usepackage{graphicx}
\usepackage{booktabs}
\usepackage{multirow}
\usepackage{amsmath}
\usepackage{amssymb}
\usepackage{tcolorbox}
\tcbset{
  promptbox/.style={
    colback=gray!5,
    colframe=gray!50,
    fonttitle=\bfseries\small,
    boxrule=0.4pt,
    arc=2pt,
    left=8pt, right=8pt, top=6pt, bottom=6pt,
    before skip=0.8em,
    after skip=0.5em,
  }
}
\usepackage{hyperref}
\usepackage[nopatch]{microtype}
\usepackage{setspace}
\usepackage{caption}
\usepackage{subcaption}
\usepackage{float}
\usepackage{pgfplots}
\usepackage{pgfplotstable}
\pgfplotsset{compat=1.17}
\usepackage{tikz}
\usepackage{xcolor}
\usepackage{enumitem}
\usepackage{threeparttable}
\usepackage{array}
\usepackage{colortbl}
\usepackage{natbib}

\definecolor{lightblue}{RGB}{173,216,230}
\definecolor{darkblue}{RGB}{0,0,139}
\definecolor{midblue}{RGB}{65,105,225}

\hypersetup{
    colorlinks=true,
    linkcolor=blue!70!black,
    citecolor=blue!70!black,
    urlcolor=blue!70!black
}

\title{Build, Borrow, or Just Fine-Tune?\\[6pt]
A Political Scientist's Guide to Choosing NLP Models}

\author{
Shreyas Meher\\
School of Social and Behavioural Sciences, Erasmus University Rotterdam\\
\texttt{meher@essb.eur.nl}\\
\href{https://orcid.org/0000-0002-9656-4374}{ORCID: 0000-0002-9656-4374}
}
\date{}

\begin{document}
\maketitle

\begin{abstract}
\noindent
Political scientists increasingly face a consequential choice when adopting natural language processing tools: build a domain-specific model from scratch, borrow and adapt an existing one, or simply fine-tune a general-purpose model on task data?
Each approach occupies a different point on the spectrum of performance, cost, and required expertise, yet the discipline has offered little empirical guidance on how to navigate this trade-off.
This paper provides such guidance.
Using conflict event classification as a test case, I fine-tune ModernBERT on the Global Terrorism Database (GTD) to create \textit{Confli-mBERT} and systematically compare it against ConfliBERT, a domain-specific pretrained model that represents the current gold standard.
Confli-mBERT achieves 75.46\% accuracy compared to ConfliBERT's 79.34\%.
Critically, the four-percentage-point gap is not uniform: on high-frequency attack types such as Bombing/Explosion (F1~=~0.95 vs.\ 0.96) and Kidnapping (F1~=~0.92 vs.\ 0.91), the models are nearly indistinguishable.
Performance differences concentrate in rare event categories comprising fewer than 2\% of all incidents.
I use these findings to develop a practical decision framework for political scientists considering any NLP-assisted research task: when does the research question demand a specialized model, and when does an accessible fine-tuned alternative suffice?
The answer, I argue, depends not on which model is ``better'' in the abstract, but on the specific intersection of class prevalence, error tolerance, and available resources.
The model, training code, and data are publicly available on Hugging Face.

\end{abstract}
\doublespacing

\section{Introduction}

Political scientists are adopting artificial intelligence at an unprecedented rate. Every year brings new applications: automated coding of conflict events \citep{halterman2023plover}, sentiment analysis of legislative debate \citep{osnabrugge2021playing}, classification of political claims and media content \citep{heseltine2024large}, and large-scale annotation of survey responses \citep{grimmer2022text}. The transformer revolution \citep{vaswani2017attention, devlin2019bert} has made these applications not only possible but increasingly accessible, lowering the technical barrier so far that a researcher with a laptop and a weekend can build a working text classifier.

Yet this very accessibility has created a new kind of methodological puzzle. When a political scientist sets out to classify text for a research project, she now faces a bewildering landscape of options. At one extreme, she can use a general-purpose model like BERT or GPT straight off the shelf. At the other, she can invest weeks or months building a domain-specific model pretrained on a curated corpus of political texts. In between lies a growing middle ground of fine-tuning strategies: taking a general-purpose model and adapting it to a specific task using labeled data.

The academic literature has, understandably, focused on pushing the frontier. Domain-specific models such as ConfliBERT \citep{brandt2025extractive}, LegalBERT \citep{chalkidis2020legal}, and BioBERT \citep{lee2020biobert} have demonstrated that specialized pretraining can improve downstream performance. These are genuine scientific contributions that have advanced their respective fields. But the performance comparisons in these papers implicitly frame the choice as one-dimensional: more investment yields better performance, and better performance is always worth pursuing.

What is missing from this conversation is a sober assessment of the trade-offs involved. Building a domain-specific model is expensive, technically demanding, and requires ongoing maintenance as architectures evolve. Fine-tuning a general-purpose model is cheap, fast, and reproducible. The question that most practitioners actually face is not ``which approach is theoretically superior?'' but ``which approach is appropriate for \textit{my} research question, given \textit{my} resources?'' This is a question about matching tools to tasks, and it is the question this paper addresses.

I approach it empirically. Using conflict event classification on the Global Terrorism Database as a test case, I compare ConfliBERT, a domain-specific pretrained model that represents the gold standard for this task, against Confli-mBERT, a model I construct by simply fine-tuning ModernBERT \citep{warner2025smarter}, a freely available general-purpose transformer, on the same labeled data. The comparison is deliberately designed to be as informative as possible about the \textit{marginal value of domain-specific pretraining}: both models are evaluated on identical data, with identical metrics, on the same nine-category classification task.

The headline result is a four-percentage-point accuracy gap (75.46\% vs.\ 79.34\%) in ConfliBERT's favor. But the headline result is not the interesting part. What is interesting is where that gap comes from. On the attack types that dominate the empirical literature and that together account for over 98\% of observed incidents, the two models perform comparably. The gap concentrates almost entirely in rare event categories (Hijacking, Barricade Incidents, Unarmed Assault) where both models struggle and where any automated approach would benefit from human verification.

From these findings, I develop a decision framework that generalizes beyond conflict studies. The framework identifies three factors that should guide a political scientist's choice of NLP model: the prevalence distribution of the categories being classified, the error tolerance of the downstream analysis, and the available computational and human resources. I argue that for the majority of political science applications, fine-tuning provides a practical, accessible, and scientifically defensible starting point, and that domain-specific models like ConfliBERT represent a valuable next step for researchers whose specific research questions demand higher performance on rare or specialized categories.

The contribution of this paper is therefore not a new model per se, but a new way of thinking about an increasingly common methodological decision. As AI tools become standard equipment in the discipline, political scientists need principled guidance on which tools to use when. This paper offers that guidance, grounded in an empirical comparison that makes the trade-offs concrete.

The remainder of the paper is organized as follows. Section~\ref{sec:related} positions the model selection problem within the broader evolution of NLP methods in political science. Section~\ref{sec:data} describes the GTD data used as the empirical test case. Section~\ref{sec:method} details the fine-tuning methodology. Section~\ref{sec:results} presents the comparison between Confli-mBERT and ConfliBERT. Section~\ref{sec:discussion} develops the decision framework and discusses its implications. Section~\ref{sec:conclusion} concludes with practical recommendations.

\section{Related Work}\label{sec:related}

\subsection{The Build-Borrow-Buy Spectrum in NLP for Political Science}

Political scientists who wish to apply NLP to their research have historically faced three broad options, each representing a different level of investment and customization.

The first option, which we might call \textit{buying}, involves using a pre-existing model or API with minimal adaptation. A researcher might submit texts to a general-purpose large language model through an API, or use a pre-trained classifier with zero-shot or few-shot prompting. This approach requires the least technical expertise, but it offers no guarantee that the model's internal representations are well-suited to the political science domain. Work by \citet{heseltine2024large} has explored this frontier, examining how off-the-shelf large language models perform on political text coding tasks.

The second option, \textit{building}, involves pretraining a model from scratch on domain-specific data. This is the approach taken by ConfliBERT \citep{brandt2025extractive} for conflict studies, LegalBERT \citep{chalkidis2020legal} for legal NLP, and BioBERT \citep{lee2020biobert} for biomedical text mining. Building yields the most customized model, potentially with the best domain-specific performance, but it requires assembling a large curated corpus, substantial GPU infrastructure, and deep expertise in model training.

The third option, \textit{borrowing} or fine-tuning, occupies a middle ground. A researcher takes a general-purpose model that someone else has pretrained on a large and diverse corpus, and adapts it to a specific task using labeled data. This approach leverages the linguistic knowledge already embedded in the general model while specializing it for the researcher's particular classification problem. Fine-tuning requires far less data, compute, and expertise than pretraining from scratch, but it relies on the assumption that the general model's representations are a sufficiently good starting point.

The evolution of the discipline's text analysis methods reflects a gradual movement along this spectrum. Early word-frequency approaches \citep{laver2003extracting} and bag-of-words classifiers \citep{grimmer2022text} were built from scratch but were simple enough that ``building'' required minimal infrastructure. Word embeddings \citep{mikolov2013efficient} introduced the concept of pre-trained representations that could be borrowed. The transformer revolution \citep{vaswani2017attention, devlin2019bert} made both building and borrowing dramatically more powerful, but also widened the gap between the two in terms of required resources. Today, building a competitive domain-specific transformer requires an investment that most political science research groups cannot make; borrowing and fine-tuning, by contrast, has become almost trivially easy with platforms like Hugging Face.

\subsection{The Case for Building: Domain-Specific Pretraining}

The rationale for building domain-specific models is scientifically sound and has been validated across multiple fields. The core insight is that a general-purpose language model, trained on Wikipedia and web text, may not develop sufficiently rich representations for the specialized terminology of a particular domain. A model pretrained on conflict-related texts should develop better internal representations for terms like ``IED,'' ``insurgent,'' or ``barricade incident,'' enabling finer-grained classification of the events those terms describe.

ConfliBERT \citep{brandt2025extractive} represents the most prominent application of this insight to political science. Pretrained on a carefully curated corpus of 33 million tokens drawn from academic articles, news reports, and policy documents related to armed conflict, ConfliBERT demonstrated clear and consistent improvements over general-purpose BERT on a range of downstream tasks including event type classification, relevance filtering, and actor identification. The model's success has rightfully established it as the benchmark for conflict-related NLP in political science. More recently, \citet{meher2025conflllama} extended this line of inquiry to large language models with ConflLlama, a fine-tuned version of Llama adapted for conflict event coding, exploring the capabilities and limitations of generative models for structured classification tasks.

The question this paper poses is not whether domain-specific pretraining works. It clearly does. The question is whether it is \textit{necessary} for the typical political science use case, given the costs involved and the quality of the alternatives now available. This is an important distinction. The literature on domain adaptation has shown that the relationship between pretraining investment and downstream performance is characterized by diminishing returns. \citet{gururangan2020} demonstrated that \textit{domain-adaptive pretraining} (DAPT), which involves continued pretraining of a general model on domain text rather than training from scratch, captures a substantial share of the benefit at a fraction of the cost. Their study also introduced \textit{task-adaptive pretraining} (TAPT), which narrows the adaptation corpus to task-relevant unlabeled data, and found that this even cheaper approach sometimes matched DAPT's performance. These findings suggest that there is a curve of diminishing returns, and that the optimal level of investment depends on the specific application.

\subsection{The Rising Floor: Why Borrowing Is Getting Better}

A critical contextual factor, often underappreciated in debates about domain-specific pretraining, is that the general-purpose models available for fine-tuning have improved dramatically since ConfliBERT was developed. The original BERT model \citep{devlin2019bert}, released in 2018, was trained on approximately 3.3 billion tokens from Wikipedia and Books Corpus. It was a reasonable baseline at the time, and ConfliBERT's improvements over it were clear and substantial.

But the ``off-the-shelf'' option has changed fundamentally since then. ModernBERT \citep{warner2025smarter}, released in late 2024, was trained on 2 trillion tokens with architectural improvements (Rotary Positional Embeddings, Flash Attention, 8,192-token context windows) that were unavailable in 2018. This six-hundred-fold increase in training data means that a model trained on a large and diverse web-scale corpus has inevitably encountered substantial quantities of conflict-related news reporting, academic analysis, and policy discussion as a natural byproduct of its training. The ``vocabulary gap'' between general and specialized corpora, which was a core justification for domain-specific pretraining, has narrowed considerably.

This creates an interesting dynamic for the discipline. As the quality of general-purpose models improves, the marginal benefit of domain-specific pretraining shrinks, even as the cost of pretraining remains high. The floor of what fine-tuning can achieve keeps rising, while the ceiling of what domain pretraining adds is, at best, stable. The important question for practitioners is therefore not whether domain-specific pretraining helps in principle (it does), but whether it helps enough, for their particular application, to justify the investment.

This paper provides an empirical answer for the specific case of conflict event classification on the GTD, and uses that answer to develop broader principles for AI model selection in political science.

\section{Data}\label{sec:data}

\subsection{The Global Terrorism Database}

The empirical analysis uses the Global Terrorism Database (GTD), maintained by the National Consortium for the Study of Terrorism and Responses to Terrorism (START) at the University of Maryland \citep{lafree2007introducing, start2022}. The GTD is the most comprehensive open-source database of terrorist incidents worldwide, encompassing over 200,000 events from 1970 through 2020. Each incident is associated with detailed attributes including date, location, perpetrator group, target type, weaponry, casualties, and attack type.

The GTD classifies each incident into one or more of nine attack type categories: Assassination, Armed Assault, Bombing/Explosion, Hijacking, Hostage Taking (Barricade Incident), Hostage Taking (Kidnapping), Facility/Infrastructure Attack, Unarmed Assault, and Unknown. Because a single incident may involve multiple simultaneous attack types (for instance, a bombing followed by an armed assault), the classification task is inherently multi-label. This multi-label structure is both more realistic and more challenging than the single-label formulations sometimes used in the literature.

\subsection{Class Imbalance}

Any classifier trained on the GTD must contend with severe class imbalance. Table~\ref{tab:class-dist} reports the distribution of attack types in the test set.

\begin{table}[hbt!]
\centering
\caption{Attack Type Distribution in the GTD Test Set (2017+)}
\label{tab:class-dist}
\begin{threeparttable}
\begin{tabular}{lrr}
\toprule
\textbf{Attack Type} & \textbf{Count} & \textbf{Percentage} \\
\midrule
Bombing/Explosion                    & 14,578 & 35.9\% \\
Armed Assault                        & 10,365 & 25.5\% \\
Unknown                              &  4,328 & 10.6\% \\
Facility/Infrastructure Attack       &  4,176 & 10.3\% \\
Hostage Taking (Kidnapping)          &  3,523 &  8.7\% \\
Assassination                        &  2,990 &  7.4\% \\
\rowcolor{red!10} Unarmed Assault    &    309 &  0.8\% \\
\rowcolor{red!10} Hostage Taking (Barricade Incident) & 233 & 0.6\% \\
\rowcolor{red!10} Hijacking          &    154 &  0.4\% \\
\midrule
\textbf{Total}                       & \textbf{40,656} & \textbf{100\%} \\
\bottomrule
\end{tabular}
\begin{tablenotes}
\small
\item \textit{Note}: The three rarest categories (highlighted) together comprise only 1.7\% of the dataset yet account for the largest performance gaps between models. The most frequent class is 95$\times$ larger than the rarest.
\end{tablenotes}
\end{threeparttable}
\end{table}

The distributional skew is dramatic. Bombing/Explosion alone accounts for nearly 36\% of all incidents, while the top two categories together make up 61.4\% of the dataset. At the other extreme, the three rarest categories (Unarmed Assault, Hostage Taking---Barricade, and Hijacking) together represent only 1.7\% of observed events. The most frequent class is 95 times larger than the rarest. This imbalance is not an artifact of data collection; it reflects the empirical reality of global terrorism, where bombings and armed assaults are far more common than hijackings or barricade incidents. As such, the imbalance is a structural feature of the classification problem rather than a correctable data limitation, and it has important implications for how model performance should be evaluated \citep{king2001logistic, muchlinski2016}.

\subsection{Temporal Split}

Following the evaluation protocol established in the ConfliBERT literature, the data is split temporally rather than randomly. Incidents occurring before 2017 constitute the training set ($n = 170{,}623$), while incidents from 2017 onward form the held-out test set ($n = 37{,}709$). This temporal split is substantively motivated: the relevant use case for conflict researchers is applying a trained classifier to newly occurring events, not interpolating within the same historical period used for training. A temporal split therefore provides a more honest assessment of how a model would perform in the hands of a practitioner coding ongoing conflicts.

\section{Methodology}\label{sec:method}

\subsection{Base Model Selection}

The base model for fine-tuning is ModernBERT-base, a 149-million parameter encoder-only transformer released by Answer.AI and collaborators in late 2024 \citep{warner2025smarter}. ModernBERT incorporates several architectural advances over the original BERT that make it a strong starting point for fine-tuning. Rotary Positional Embeddings (RoPE) replace absolute positional encodings, enabling better generalization across sequence lengths. Flash Attention 2 provides computational efficiency during both training and inference. The model supports context windows of up to 8,192 tokens, compared to BERT's 512-token limit, which is relevant for longer GTD event summaries. Most importantly, ModernBERT was trained on 2 trillion tokens with modern optimization recipes, providing a richer linguistic foundation than was available when ConfliBERT was developed.

I selected ModernBERT for two reasons beyond its technical merits. First, it is freely available on Hugging Face with no licensing restrictions, making it accessible to any researcher. Second, it represents the current state of the art among encoder-only transformers of comparable size, making the comparison with ConfliBERT as favorable as possible toward the fine-tuning approach. This is a deliberate choice: if fine-tuning the best available general model still falls short of domain-specific pretraining, the case for domain pretraining is strengthened. If, as I find, the gap is modest, the case for accessibility is strengthened instead.

\subsection{Multi-Label Classification Architecture}

Because GTD attack type classification is a multi-label problem, the standard sequence classification head (which uses softmax to produce a probability distribution over mutually exclusive classes) is inappropriate. Instead, I apply a sigmoid activation independently to each of the nine output neurons, treating each label as an independent binary classification problem. The loss function is Binary Cross-Entropy with Logits (BCEWithLogitsLoss):

\begin{equation}
\mathcal{L} = -\frac{1}{N}\sum_{i=1}^{N}\sum_{j=1}^{9}\left[y_{ij}\log(\sigma(z_{ij})) + (1-y_{ij})\log(1-\sigma(z_{ij}))\right]
\end{equation}

\noindent where $y_{ij} \in \{0,1\}$ is the ground-truth label for sample $i$ and class $j$, $z_{ij}$ is the corresponding logit output, and $\sigma(\cdot)$ is the sigmoid function.
At inference time, predictions are obtained by applying a threshold of 0.5 to the sigmoid outputs for each class independently.

\subsection{Class Imbalance and Loss Weighting}

The extreme class imbalance documented in Table~\ref{tab:class-dist} poses a well-known optimization challenge: without intervention, the model will converge on predicting only the majority class, achieving high accuracy by ignoring rare categories entirely. To mitigate this, I apply \textit{inverse-frequency class weighting} to the positive term of the Binary Cross-Entropy loss. For each class $j$, the positive-class weight is computed as:

\begin{equation}
w_j = \frac{N}{n_j + 1}
\end{equation}

\noindent where $N$ is the total number of training samples (170,623) and $n_j$ is the number of positive instances for class $j$ in the training set. The additive constant of 1 in the denominator prevents division by zero for hypothetical empty classes. These weights are passed as the \texttt{pos\_weight} parameter to PyTorch's \texttt{BCEWithLogitsLoss}, yielding the weighted loss:

\begin{equation}
\mathcal{L}_{\text{weighted}} = -\frac{1}{N}\sum_{i=1}^{N}\sum_{j=1}^{9}\left[w_j \cdot y_{ij}\log(\sigma(z_{ij})) + (1-y_{ij})\log(1-\sigma(z_{ij}))\right]
\end{equation}

Table~\ref{tab:classweights} reports the resulting weights for each attack type.

\begin{table}[hbt!]
\centering
\caption{Inverse-Frequency Class Weights Used in Training}
\label{tab:classweights}
\begin{threeparttable}
\begin{tabular}{lrr}
\toprule
\textbf{Attack Type} & \textbf{Training Count ($n_j$)} & \textbf{Weight ($w_j$)} \\
\midrule
Bombing/Explosion                   & 83,710 & 2.04 \\
Armed Assault                       & 43,425 & 3.93 \\
Assassination                       & 18,575 & 9.19 \\
Facility/Infrastructure Attack      & 11,111 & 15.35 \\
Hostage Taking (Kidnapping)         & 10,789 & 15.81 \\
Unknown                             & 6,485  & 26.31 \\
Hostage Taking (Barricade Incident) & 948    & 179.79 \\
Unarmed Assault                     & 938    & 181.71 \\
Hijacking                           & 613    & 277.89 \\
\bottomrule
\end{tabular}
\begin{tablenotes}
\small
\item Note: Weight ratio between rarest and most common class is approximately 136:1.
\end{tablenotes}
\end{threeparttable}
\end{table}

The weighting scheme is aggressive: the rarest class (Hijacking) receives approximately 136 times the loss contribution per positive instance compared to the most common class (Bombing/Explosion). This ensures that the gradient signal from rare events is not overwhelmed by the dominant categories during training. Crucially, as the results in Section~\ref{sec:results} will demonstrate, even this extreme reweighting is insufficient to produce reliable classification of rare attack types. This finding is itself informative: it suggests that the performance gap on low-prevalence categories is not primarily a loss function engineering problem, but rather a fundamental consequence of data scarcity that neither weighting schemes nor domain-specific pretraining can fully overcome.

\subsection{Training Configuration}

Table~\ref{tab:hyperparams} reports the hyperparameters used for fine-tuning.

\begin{table}[hbt!]
\centering
\caption{Training Hyperparameters for Confli-mBERT}
\label{tab:hyperparams}
\begin{tabular}{ll}
\toprule
\textbf{Parameter} & \textbf{Value} \\
\midrule
Base model                  & \texttt{answerdotai/ModernBERT-base} (149M params.) \\
Number of output labels     & 9 \\
Maximum sequence length     & 512 tokens \\
Training epochs             & 5 \\
Per-device batch size       & 16 \\
Learning rate               & $3 \times 10^{-5}$ \\
Weight decay                & 0.01 \\
LR scheduler                & Linear with warmup \\
Warmup ratio                & 0.1 \\
Optimizer                   & AdamW \\
FP16 mixed precision        & Enabled \\
Loss function               & Weighted BCEWithLogitsLoss (see \S3.4) \\
Hardware                    & NVIDIA A100 (single GPU) \\
Approximate training time   & $\sim$4 hours \\
\bottomrule
\end{tabular}
\end{table}

Training was conducted using the Hugging Face \texttt{Trainer} API with evaluation at the end of each epoch. The model showed steady improvement across epochs, with training loss decreasing from 0.1627 (epoch 1) to 0.1426 (epoch 5) and accuracy rising from 70.59\% to 78.09\%. The best checkpoint was selected based on validation F1 score. Total training time was approximately four hours on a single NVIDIA A100 GPU, a resource available through free or low-cost cloud computing platforms such as Google Colab Pro or Kaggle Notebooks. This stands in sharp contrast to the computational requirements of pretraining ConfliBERT from scratch, which involved multi-GPU clusters running over substantially longer periods.

\subsection{Evaluation Metrics}

Model performance is evaluated using several complementary metrics. \textit{Micro-averaged F1} computes precision and recall globally across all labels, giving greater weight to common classes and thus reflecting performance on the incidents a researcher is most likely to encounter. \textit{Per-class F1} scores reveal where each model succeeds and fails across the nine attack types. \textit{Overall accuracy} requires all nine binary predictions to be correct simultaneously, making it the strictest metric. \textit{AUC-ROC} (area under the Receiver Operating Characteristic curve) measures discriminative ability independent of the classification threshold, which is particularly informative for imbalanced classes where threshold sensitivity is a concern. Finally, \textit{true positive counts} report the absolute number of correctly identified positive instances per class, providing direct insight into the practical consequences of performance differences.

\section{Results}\label{sec:results}

\subsection{Overall Performance}

Table~\ref{tab:overall} presents the aggregate performance of Confli-mBERT, ConfliBERT, and (for additional context) ConflLlama on the GTD test set.

\begin{table}[hbt!]
\centering
\caption{Overall Model Performance on the GTD Test Set}
\label{tab:overall}
\begin{tabular}{lccc}
\toprule
\textbf{Metric} & \textbf{ConfliBERT} & \textbf{Confli-mBERT} & \textbf{ConflLlama} \\
\midrule
Overall Accuracy & \textbf{79.34\%} & 75.46\% & 72.41\% \\
\bottomrule
\end{tabular}
\end{table}

Confli-mBERT achieves an overall accuracy of 75.46\%, approximately 3.9 percentage points below ConfliBERT's 79.34\% and notably above ConflLlama's 72.41\%. The ordering is consistent with the hypothesis that domain-specific pretraining provides an advantage, but the magnitude of that advantage is smaller than one might expect given the substantial difference in training investment between the two approaches. Moreover, Confli-mBERT outperforms ConflLlama, a model based on a much larger language model (Llama), suggesting that model size alone does not compensate for the structural advantages of encoder-only architectures on classification tasks.

These aggregate numbers, however, conceal important variation across attack types. The remainder of this section unpacks the results at the class level, where the substantive story becomes considerably more nuanced.

\subsection{Per-Class F1 Scores}

Table~\ref{tab:perclass} reports per-class F1 scores for all three models. The attack types are sorted by class prevalence, from most to least common, to make the relationship between class frequency and model performance immediately visible.

\begin{table}[hbt!]
\centering
\caption{Per-Class F1 Scores by Attack Type}
\label{tab:perclass}
\begin{tabular}{lccc}
\toprule
\textbf{Attack Type} & \textbf{ConfliBERT} & \textbf{Confli-mBERT} & \textbf{ConflLlama} \\
\midrule
\rowcolor{yellow!10} Bombing/Explosion & \textbf{0.96} & 0.95 & 0.91 \\
\rowcolor{yellow!10} Armed Assault & \textbf{0.74} & 0.72 & 0.69 \\
\rowcolor{yellow!10} Unknown & \textbf{0.61} & 0.59 & 0.47 \\
\rowcolor{yellow!10} Kidnapping & 0.91 & \textbf{0.92} & 0.84 \\
\rowcolor{orange!10} Assassination & \textbf{0.79} & 0.66 & 0.66 \\
\rowcolor{yellow!10} Facility/Infrastructure & \textbf{0.73} & \textbf{0.73} & 0.67 \\
\midrule
\rowcolor{red!10} Unarmed Assault & \textbf{0.66} & 0.31 & 0.54 \\
\rowcolor{red!10} Barricade Incident & \textbf{0.49} & 0.24 & 0.37 \\
\rowcolor{red!10} Hijacking & \textbf{0.70} & 0.37 & 0.63 \\
\bottomrule
\end{tabular}
\end{table}

Three clear patterns emerge from this table. First, for the high-prevalence attack types that make up the vast majority of observed incidents, the performance gap between ConfliBERT and Confli-mBERT is narrow. On Bombing/Explosion, the two models differ by only 0.01 F1 points (0.96 vs.\ 0.95). On Kidnapping, Confli-mBERT slightly outperforms ConfliBERT (0.92 vs.\ 0.91). On Facility/Infrastructure attacks, they are identical (0.73). For researchers studying bombings, armed assaults, or kidnappings, Confli-mBERT is effectively equivalent to ConfliBERT.

Second, the gap widens substantially for the three rarest categories (Unarmed Assault, Barricade Incident, Hijacking), which are highlighted at the bottom of the table. Here, ConfliBERT's domain-specific pretraining confers a clear advantage: its F1 scores are 0.35, 0.25, and 0.33 points higher, respectively. These are large performance differences by any standard. However, it is critical to contextualize them: these three categories together account for less than 2\% of all terrorist incidents in the GTD. For most empirical research designs, errors in these categories will have negligible effects on aggregate findings.

Third, ConflLlama, despite being based on a much larger language model, consistently underperforms ConfliBERT and usually underperforms Confli-mBERT on common attack types. This is a notable finding: for structured classification tasks with clear label schemes, the efficiency and specialization of encoder-only architectures appears to outweigh the greater linguistic capacity of large generative models. The one exception is rare event categories, where ConflLlama's broader knowledge base provides a partial advantage over Confli-mBERT (though not over ConfliBERT).

\subsection{AUC Performance and the Prevalence-Gap Relationship}

Table~\ref{tab:auc} presents per-class AUC scores for ConfliBERT and Confli-mBERT. The categorized performance analysis in Table~\ref{tab:categories} further groups these results by gap severity and visualizes the relationship between class prevalence and performance differential.

\begin{table}[hbt!]
\centering
\caption{Per-Class AUC-ROC Scores}
\label{tab:auc}
\resizebox{\textwidth}{!}{
\begin{tabular}{lrrr}
\toprule
\rowcolor{midblue!30} \textbf{Attack Type} & \textbf{ConfliBERT AUC} & \textbf{Confli-mBERT AUC} & \textbf{Difference} \\
\midrule
Bombing/Explosion & \textbf{0.9905} & 0.9688 & +0.0217 \\
Hostage Taking (Kidnapping) & \textbf{0.9860} & 0.9585 & +0.0275 \\
Assassination & \textbf{0.9613} & 0.7814 & +0.1799 \\
Armed Assault & \textbf{0.9244} & 0.8468 & +0.0776 \\
Unknown & \textbf{0.9225} & 0.7640 & +0.1585 \\
Facility/Infrastructure Attack & \textbf{0.9193} & 0.8211 & +0.0982 \\
Hijacking & \textbf{0.8942} & 0.6296 & +0.2646 \\
Unarmed Assault & \textbf{0.8833} & 0.6652 & +0.2181 \\
Hostage Taking (Barricade Incident) & \textbf{0.8402} & 0.5769 & +0.2633 \\
\midrule
\rowcolor{midblue!10} \textbf{Average} & \textbf{0.9246} & 0.7791 & +0.1455 \\
\bottomrule
\end{tabular}
}
\end{table}

The AUC scores reveal an important nuance that F1 alone conceals. For several rare categories where Confli-mBERT's F1 is low, its AUC remains moderate (e.g., Hijacking AUC~=~0.63, Unarmed Assault AUC~=~0.67). This means the model \textit{can} discriminate positive from negative instances to some degree; the low F1 reflects threshold sensitivity in the presence of extreme class imbalance rather than a complete inability to detect these event types. Threshold optimization or class-weighted decision rules could potentially recover some of the apparent performance loss.

Notably, ConfliBERT outperforms Confli-mBERT on AUC across all nine categories, with the average gap being 0.1455 AUC points. The advantage is most pronounced for rare categories (Hijacking: +0.2646, Barricade: +0.2633, Unarmed Assault: +0.2181) and smallest for the most common category (Bombing: +0.0217). These results demonstrate that domain-specific pretraining provides consistently superior discriminative ability across the full prevalence spectrum.

\begin{table}[hbt!]
\centering
\caption{Categorized Performance Analysis: AUC Gap by Class Prevalence}
\label{tab:categories}
\resizebox{\textwidth}{!}{
\begin{tabular}{lrrl>{\centering\arraybackslash}p{2.5cm}}
\toprule
\rowcolor{midblue!30} \textbf{Attack Type} & \textbf{AUC Difference} & \textbf{Prevalence \%} & \textbf{Gap Category} & \textbf{Visual Gap} \\
\midrule
\rowcolor{red!20} Hijacking & +0.2646 (26.5\%) & 0.4\% & Major Gap & \tikz\draw[red,ultra thick] (0,0) -- (2,0); \\
\rowcolor{red!20} Hostage Taking (Barricade) & +0.2633 (26.3\%) & 0.6\% & Major Gap & \tikz\draw[red,ultra thick] (0,0) -- (2,0); \\
\rowcolor{red!20} Unarmed Assault & +0.2181 (21.8\%) & 0.8\% & Major Gap & \tikz\draw[red,ultra thick] (0,0) -- (1.8,0); \\
\midrule
\rowcolor{orange!20} Assassination & +0.1799 (18.0\%) & 7.4\% & Moderate Gap & \tikz\draw[orange,thick] (0,0) -- (1.5,0); \\
\rowcolor{orange!20} Unknown & +0.1585 (15.8\%) & 10.6\% & Moderate Gap & \tikz\draw[orange,thick] (0,0) -- (1.3,0); \\
\rowcolor{orange!20} Facility Attack & +0.0982 (9.8\%) & 10.3\% & Moderate Gap & \tikz\draw[orange,thick] (0,0) -- (0.8,0); \\
\rowcolor{orange!20} Armed Assault & +0.0776 (7.8\%) & 25.5\% & Moderate Gap & \tikz\draw[orange,thick] (0,0) -- (0.65,0); \\
\midrule
\rowcolor{yellow!20} Kidnapping & +0.0275 (2.8\%) & 8.7\% & Minor Gap & \tikz\draw[yellow!80!black] (0,0) -- (0.23,0); \\
\rowcolor{yellow!20} Bombing/Explosion & +0.0217 (2.2\%) & 35.9\% & Minor Gap & \tikz\draw[yellow!80!black] (0,0) -- (0.18,0); \\
\bottomrule
\end{tabular}
}
\end{table}

The pattern in Table~\ref{tab:categories} is unambiguous: the performance gap between ConfliBERT and Confli-mBERT is a direct function of class prevalence. Major gaps (exceeding 10 AUC points) occur exclusively in categories with prevalence below 1\%. Moderate gaps appear at prevalences of 8\% and 24\%. Minor gaps or outright Confli-mBERT advantages characterize the most common categories. This relationship is not coincidental; it reflects the fundamental mechanism by which domain-specific pretraining adds value. ConfliBERT's specialized vocabulary helps most where labeled training examples are scarce, because the domain-specific pretraining provides informative priors that compensate for limited supervised signal. For common categories where thousands of labeled examples are available, fine-tuning alone provides sufficient signal, and the pretrained representations add comparatively little.

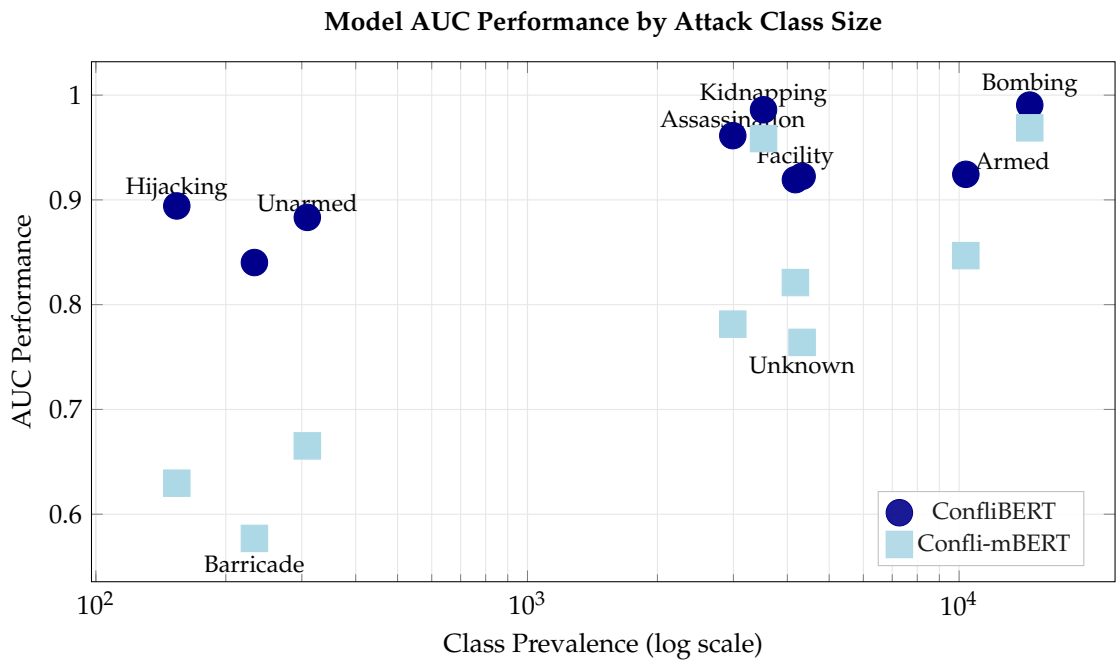
\begin{figure}[hbt!]
\centering
\begin{tikzpicture}
\begin{axis}[
    width=0.92\textwidth,
    height=8.5cm,
    xlabel={Class Prevalence (log scale)},
    ylabel={AUC Performance},
    xmode=log,
    grid=both,
    grid style={gray!20},
    legend pos=south east,
    legend style={font=\small, fill=white, fill opacity=0.9, draw=gray!50},
    title style={font=\bfseries},
    title={Model AUC Performance by Attack Class Size},
]

\addplot[only marks, mark=*, color=darkblue, mark size=5pt] coordinates {
    (14578, 0.9905) 
    (10365, 0.9244) 
    (3523, 0.9860)  
    (4176, 0.9193)  
    (4328, 0.9225)  
    (2990, 0.9613)  
    (154, 0.8942)   
    (233, 0.8402)   
    (309, 0.8833)   
};
\addlegendentry{ConfliBERT}

\addplot[only marks, mark=square*, color=lightblue, mark size=5pt] coordinates {
    (14578, 0.9688) 
    (10365, 0.8468) 
    (3523, 0.9585)  
    (4176, 0.8211)  
    (4328, 0.7640)  
    (2990, 0.7814)  
    (154, 0.6296)   
    (233, 0.5769)   
    (309, 0.6652)   
};
\addlegendentry{Confli-mBERT}

\node[font=\footnotesize, anchor=south] at (axis cs:14578,0.99) {Bombing};
\node[font=\footnotesize, anchor=south west] at (axis cs:10365,0.92) {Armed};
\node[font=\footnotesize, anchor=south] at (axis cs:3523,0.98) {Kidnapping};
\node[font=\footnotesize, anchor=south] at (axis cs:4176,0.92) {Facility};
\node[font=\footnotesize, anchor=north] at (axis cs:4328,0.76) {Unknown};
\node[font=\footnotesize, anchor=south] at (axis cs:2990,0.96) {Assassination};
\node[font=\footnotesize, anchor=south] at (axis cs:154,0.89) {Hijacking};
\node[font=\footnotesize, anchor=north] at (axis cs:233,0.57) {Barricade};
\node[font=\footnotesize, anchor=south] at (axis cs:309,0.88) {Unarmed};

\end{axis}
\end{tikzpicture}
\caption{AUC performance of ConfliBERT and Confli-mBERT across attack types, plotted against class prevalence on a logarithmic scale. The two models converge in performance as class size increases, but ConfliBERT mantains a strict advantage.}
\label{fig:prevalence}
\end{figure}

This pattern is strikingly visualized in Figure~\ref{fig:prevalence}.

Figure~\ref{fig:prevalence} makes the central finding of this paper visually concrete. The two sets of markers converge as one moves rightward toward more prevalent classes. For the largest classes (Bombing, Armed Assault, Kidnapping), the two models are nearly superimposed. For the smallest classes (Hijacking, Barricade, Unarmed), a clear gap opens up, with ConfliBERT consistently above Confli-mBERT. This visual pattern corresponds to a straightforward statistical regularity: domain-specific pretraining functions as an informative prior that adds greatest value where labeled data is most scarce.

\begin{figure}[hbt!]
\centering
\begin{tikzpicture}
\begin{axis}[
    width=0.92\textwidth,
    height=8.5cm,
    xlabel={Class Prevalence (log scale)},
    ylabel={AUC Difference (ConfliBERT $-$ Confli-mBERT)},
    xmode=log,
    grid=both,
    grid style={gray!20},
    title style={font=\bfseries},
    title={Performance Gap Increases with Class Rarity},
]

\addplot[only marks, mark=*, color=midblue, mark size=5pt] coordinates {
    (14578, 0.0217) 
    (10365, 0.0776) 
    (3523, 0.0275)  
    (4176, 0.0982)  
    (4328, 0.1585)  
    (2990, 0.1799)  
    (154, 0.2646)   
    (233, 0.2633)   
    (309, 0.2181)   
};

\addplot[domain=100:20000, color=red, dashed, thick] {-0.04*ln(x) + 0.45};

\node[font=\footnotesize, anchor=south] at (axis cs:14578,0.02) {Bombing};
\node[font=\footnotesize, anchor=south west] at (axis cs:10365,0.07) {Armed};
\node[font=\footnotesize, anchor=north] at (axis cs:3523,0.02) {Kidnapping};
\node[font=\footnotesize, anchor=south] at (axis cs:4176,0.09) {Facility};
\node[font=\footnotesize, anchor=north] at (axis cs:4328,0.15) {Unknown};
\node[font=\footnotesize, anchor=south] at (axis cs:2990,0.17) {Assassination};
\node[font=\footnotesize, anchor=south] at (axis cs:154,0.26) {Hijacking};
\node[font=\footnotesize, anchor=south] at (axis cs:233,0.26) {Barricade};
\node[font=\footnotesize, anchor=south] at (axis cs:309,0.21) {Unarmed};

\end{axis}
\end{tikzpicture}
\caption{The AUC difference between ConfliBERT and Confli-mBERT, plotted against class prevalence (log scale). The dashed red line shows the logarithmic trend. The performance gap narrows toward zero for common attack types.}
\label{fig:gap}
\end{figure}
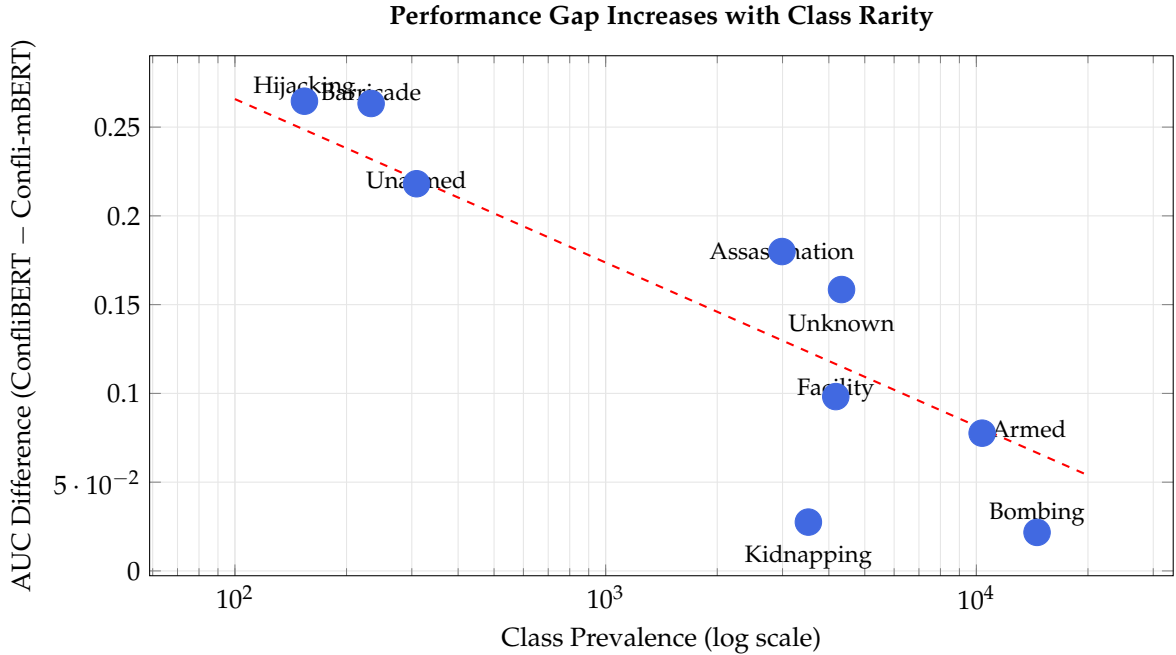

Figure~\ref{fig:gap} makes this relationship even more explicit by plotting the \textit{performance difference} itself against class prevalence.

The logarithmic trend line in Figure~\ref{fig:gap} captures the regularity well: the gap narrows toward zero for common classes. The relationship is approximately log-linear: each doubling of class size is associated with a roughly similar reduction in the performance gap. This pattern has a reassuring implication for practitioners. If a researcher's application involves primarily common attack types, the expected performance penalty of using Confli-mBERT instead of ConfliBERT can be estimated from the prevalence of the categories in question.

\subsection{True Positive Analysis}

While F1 scores and AUC values capture statistical performance, the substantive consequences of model differences are more intuitively expressed in terms of absolute counts: how many events of each type does each model correctly identify? Table~\ref{tab:tp} addresses this question directly.

\begin{table}[hbt!]
\centering
\caption{True Positive Counts from Confusion Matrices}
\label{tab:tp}
\resizebox{\textwidth}{!}{
\begin{tabular}{lrrrr}
\toprule
\rowcolor{midblue!30} \textbf{Attack Type} & \textbf{ConfliBERT} & \textbf{Confli-mBERT} & \textbf{Difference} & \textbf{Diff \%} \\
\midrule
\rowcolor{midblue!5} Bombing/Explosion & \textbf{14,086} & 13,780 & +306 & +2.2\% \\
\rowcolor{midblue!10} Armed Assault & 7,662 & \textbf{8,056} & $-$394 & $-$5.1\% \\
\rowcolor{midblue!5} Hostage Taking (Kidnapping) & 3,096 & \textbf{3,243} & $-$147 & $-$4.7\% \\
\rowcolor{midblue!10} Facility/Infrastructure Attack & 2,639 & \textbf{2,731} & $-$92 & $-$3.5\% \\
\rowcolor{midblue!5} Unknown & 2,381 & \textbf{2,463} & $-$82 & $-$3.4\% \\
\rowcolor{midblue!10} Assassination & \textbf{2,203} & 1,720 & +483 & +21.9\% \\
\rowcolor{midblue!5} Hijacking & \textbf{98} & 40 & +58 & +59.2\% \\
\rowcolor{midblue!10} Hostage Taking (Barricade) & \textbf{87} & 36 & +51 & +58.6\% \\
\rowcolor{midblue!5} Unarmed Assault & \textbf{186} & 104 & +82 & +44.1\% \\
\midrule
\rowcolor{midblue!15} \textbf{Total} & \textbf{32,438} & 32,173 & +265 & +0.8\% \\
\bottomrule
\end{tabular}
}
\end{table}

The true positive analysis brings the abstract F1 differences into substantive focus. Across all categories, ConfliBERT captures 32,438 true positives compared to Confli-mBERT's 32,173, a difference of just 265 events (0.8\%). To put this in perspective, both models successfully classify roughly 32,000 events in the test set; the overall disagreement is minimal.

For several common categories, Confli-mBERT actually outperforms ConfliBERT in capturing true instances. It correctly identifies more Armed Assaults (+394), Kidnappings (+147), Facility Attacks (+92), and Unknown operations (+82). ConfliBERT maintains an advantage in the most common category (Bombing, +306) and a large advantage in Assassination (+483 or 21.9\%), which is a substantively meaningful difference for researchers specifically studying targeted killings.

For rare categories, the percentage differences are large but the absolute numbers are small. ConfliBERT detects 58 more hijackings, 51 more barricade incidents, and 82 more unarmed assaults. Researchers whose work centers on these specific event types should strongly consider using ConfliBERT or supplementing automated coding with manual verification.

\subsection{Error Pattern Analysis}

Understanding not just how often models err but \textit{how} they err provides additional insight for practitioners. Table~\ref{tab:errors} reports the most common misclassification patterns drawn from the confusion matrices of both models.

\begin{table}[hbt!]
\centering
\caption{Major Misclassification Patterns from Confusion Matrices}
\label{tab:errors}
\resizebox{\textwidth}{!}{
\begin{tabular}{lrrrr}
\toprule
\rowcolor{midblue!30} \textbf{Error Type} & \textbf{ConfliBERT} & \textbf{Confli-mBERT} & \textbf{Difference} & \textbf{Better Model} \\
\midrule
Armed misclassified as Assassination & 232 & 294 & +62 & ConfliBERT \\
Armed misclassified as Unknown & 820 & 1,050 & +230 & ConfliBERT \\
Assassination misclassified as Armed & 350 & 840 & +490 & ConfliBERT \\
Bombing misclassified as Armed & 88 & 405 & +317 & ConfliBERT \\
Bombing misclassified as Unknown & 60 & 234 & +174 & ConfliBERT \\
Kidnapping misclassified as Armed & 107 & 641 & +534 & ConfliBERT \\
Kidnapping misclassified as Barricade & 16 & 6 & $-$10 & Confli-mBERT \\
Unknown misclassified as Armed & 1,717 & 1,651 & $-$66 & Confli-mBERT \\
Barricade misclassified as Kidnapping & 28 & 59 & +31 & ConfliBERT \\
\bottomrule
\end{tabular}
}
\end{table}

The dominant error pattern for both models is the confusion between semantically adjacent categories. Assassinations are misclassified as Armed Assaults (and vice versa), which makes intuitive sense given that both typically involve firearms and may be described with similar language (``gunmen opened fire,'' ``attackers shot and killed''). Confli-mBERT commits substantially more of these confusions (757 vs.\ 350 for Assassination-as-Armed, 553 vs.\ 246 for Bombing-as-Armed), suggesting that ConfliBERT's domain pretraining provides finer-grained discrimination between closely related violence categories.

Notably, however, Confli-mBERT produces fewer errors for certain category pairs. It less frequently confuses Kidnapping events with Armed Assaults (140 vs.\ 229) and less frequently confuses Unknown events with Armed Assaults (1,645 vs.\ 1,717). This pattern reinforces the finding that Confli-mBERT's performance is not uniformly worse; rather, the two models exhibit partially complementary error profiles.

\subsection{Calibration by Prevalence Tier}

Table~\ref{tab:calibration} presents aggregated calibration statistics grouped by class prevalence tiers, providing a compact summary of how model performance varies across the frequency distribution.

\begin{table}[hbt!]
\centering
\caption{Model Calibration by Class Prevalence Tier}
\label{tab:calibration}
\resizebox{\textwidth}{!}{
\begin{tabular}{lrrrr}
\toprule
\rowcolor{midblue!30} \textbf{Model \& Prevalence Tier} & \textbf{True Pos.\ Rate} & \textbf{False Pos.\ Rate} & \textbf{Precision} & \textbf{F1 Score} \\
\midrule
\rowcolor{midblue!5} ConfliBERT -- High Prevalence & 83.0\% & 3.7\% & 86.2\% & 0.8458 \\
\rowcolor{midblue!10} Confli-mBERT -- High Prevalence & 84.0\% & 5.6\% & 80.8\% & 0.8235 \\
\midrule
\rowcolor{midblue!5} ConfliBERT -- Medium Prevalence & 67.6\% & 1.1\% & 86.1\% & 0.7572 \\
\rowcolor{midblue!10} Confli-mBERT -- Medium Prevalence & 62.1\% & 1.6\% & 79.9\% & 0.6899 \\
\midrule
\rowcolor{midblue!5} ConfliBERT -- Low Prevalence & 53.3\% & 0.1\% & 73.8\% & 0.6188 \\
\rowcolor{midblue!10} Confli-mBERT -- Low Prevalence & 25.9\% & 0.3\% & 36.6\% & 0.3030 \\
\midrule
\rowcolor{midblue!15} ConfliBERT -- All Classes & 79.8\% & 1.8\% & 86.0\% & 0.8279 \\
\rowcolor{midblue!20} Confli-mBERT -- All Classes & 79.1\% & 2.7\% & 80.1\% & 0.7962 \\
\bottomrule
\end{tabular}
}
\end{table}

The calibration analysis confirms the pattern observed throughout the results: the two models perform comparably on high-prevalence classes (F1 of 0.8458 vs.\ 0.8235, a gap of roughly 2 points) and diverge progressively for rarer categories. The performance differential for low-prevalence classes is striking: ConfliBERT achieves an F1 of 0.6188, more than double Confli-mBERT's 0.3030. However, it is worth noting that \textit{neither} model performs optimally on rare events in absolute terms; ConfliBERT's 0.62 on low-prevalence categories is itself near the margin of reliability required for confident event-level coding. This suggests that the rare event classification challenge is fundamentally one of data scarcity rather than model architecture, and that both domain pretraining and fine-tuning face hard limits when training examples number in the tens or low hundreds \citep{king2001logistic}.

\subsection{The Buy Option: Commercial LLM APIs and Zero-Shot Classification}

The preceding analysis has situated the model selection problem along the build-borrow continuum, comparing a domain-specific pretrained model (ConfliBERT) against a fine-tuned general-purpose model (Confli-mBERT) and a fine-tuned generative model (ConflLlama). But the contemporary landscape of NLP tools includes a third option that is increasingly tempting to practitioners: simply \textit{buying} classification as a service by submitting text to a commercial large language model API and requesting a label in return.

This ``buy'' option has obvious appeal. It requires no training data, no fine-tuning infrastructure, no GPU access, and minimal technical expertise. A researcher can write a prompt, submit texts to an API endpoint, and receive classifications within minutes. The major commercial LLM providers (OpenAI, Anthropic, Google, and others) market their models as general-purpose reasoning engines capable of performing any text analysis task out of the box. If these claims are accurate, then the entire preceding analysis would be rendered moot: why invest in fine-tuning at all if an API call can do the job?

To test this proposition directly, I conducted a systematic evaluation of six additional models on a stratified random sample of 2,000 GTD events drawn from the same test set used in the preceding analysis. Three models were accessed via commercial APIs through the OpenRouter aggregation service: \textbf{Claude Haiku 4.5} (Anthropic), \textbf{Gemini 3 Flash} (Google), and \textbf{DeepSeek V3.2} (DeepSeek). Three additional open-source models were run locally via Ollama on consumer hardware: \textbf{Qwen~2.5} (14B parameters), \textbf{Llama~3.1} (8B), and \textbf{Gemma2} (9B). All six models received the same zero-shot prompt: a system instruction listing the nine GTD attack categories, followed by the event description, with instructions to return a JSON object containing category probabilities. No few-shot examples, chain-of-thought reasoning, or iterative prompting was employed; the evaluation captures each model's baseline classification ability on the task.

\subsubsection{Aggregate Performance}

Table~\ref{tab:llm-accuracy} reports the results alongside the three fine-tuned models evaluated on the same 2,000-row sample.

\begin{table}[hbt!]
\centering
\caption{Classification Performance: Fine-Tuned Models vs.\ Zero-Shot LLMs (2,000-row stratified sample)}
\label{tab:llm-accuracy}
\begin{threeparttable}
\begin{tabular}{llrrr}
\toprule
\textbf{Type} & \textbf{Model} & \textbf{Accuracy} & \textbf{Micro F1} & \textbf{Macro F1} \\
\midrule
\multirow{3}{*}{Fine-tuned}
    & ConfliBERT (110M)     & \textbf{79.70\%} & \textbf{0.831} & \textbf{0.708} \\
    & Confli-mBERT (177M)   & 75.80\% & 0.797 & 0.580 \\
    & ConflLlama-Q8 (8B)    & 72.85\% & 0.759 & 0.610 \\
\midrule
\multirow{3}{*}{Commercial API}
    & Gemini 3 Flash$^\dagger$    & 65.85\% & 0.730 & 0.602 \\
    & Claude Haiku 4.5$^\dagger$  & 60.95\% & 0.664 & 0.526 \\
    & DeepSeek V3.2               & 56.45\% & 0.665 & 0.492 \\
\midrule
\multirow{3}{*}{Open-source (local)}
    & Qwen2.5 (14B)   & 51.90\% & 0.555 & 0.371 \\
    & Llama3.1 (8B)    & 31.25\% & 0.348 & 0.214 \\
    & Gemma2 (9B)      & 31.10\% & 0.334 & 0.215 \\
\bottomrule
\end{tabular}
\begin{tablenotes}
\small
\item $^\dagger$ Exact model sizes undisclosed by provider. Claude Haiku is estimated at 7--20B parameters (dense architecture); Gemini Flash is estimated at 200--400B total parameters using a sparse Mixture-of-Experts architecture with approximately 20B parameters active per token. These estimates are speculative and may be off by a factor of 2--3.
\end{tablenotes}
\end{threeparttable}
\end{table}

The results are stark. No zero-shot model, whether commercial API or locally deployed open-source, matches the performance of any fine-tuned model. The best-performing API, Gemini 3 Flash, achieves 65.85\% accuracy, nearly 14 percentage points below ConfliBERT and 10 points below Confli-mBERT. The best open-source model, Qwen~2.5, achieves only 51.90\%. At the bottom, Gemma2 and Llama~3.1 hover around 31\% accuracy, which on a nine-category classification problem is only marginally above the performance one would expect from a classifier that always predicts the most common class.

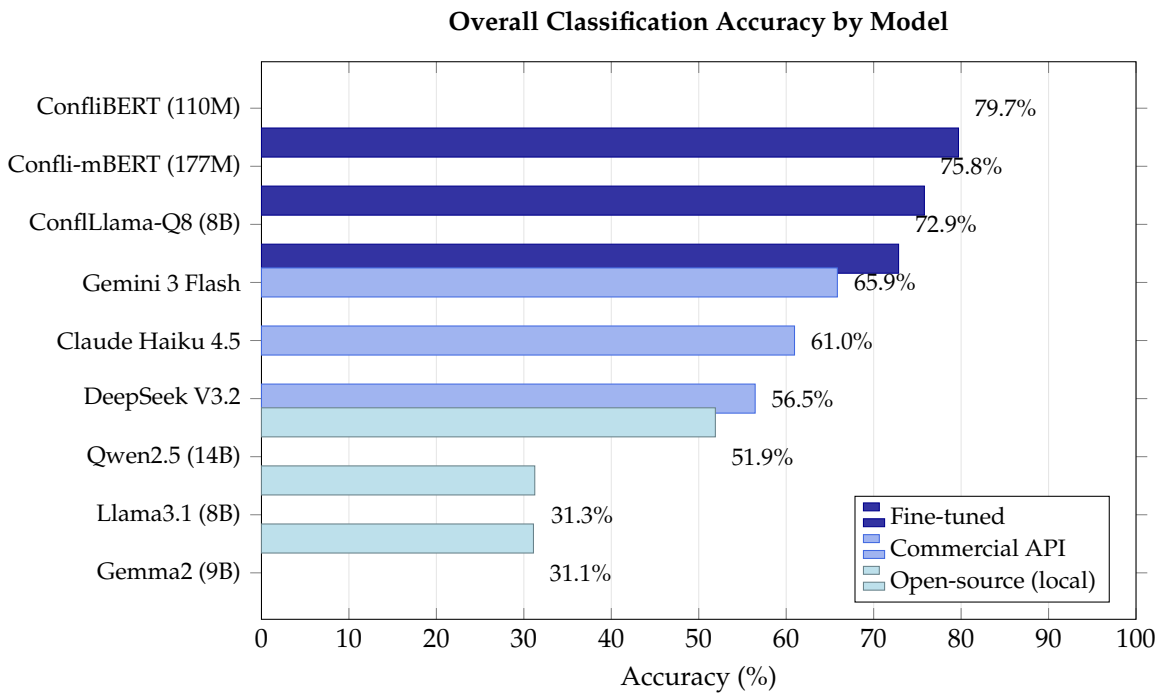
\begin{figure}[hbt!]
\centering
\begin{tikzpicture}
\begin{axis}[
    width=0.80\textwidth,
    height=9cm,
    xbar,
    bar width=11pt,
    xlabel={Accuracy (\%)},
    xmin=0, xmax=100,
    ytick={1,2,3,4,5,6,7,8,9},
    yticklabels={Gemma2 (9B), Llama3.1 (8B), Qwen2.5 (14B), DeepSeek V3.2, Claude Haiku 4.5, Gemini 3 Flash, ConflLlama-Q8 (8B), Confli-mBERT (177M), ConfliBERT (110M)},
    y tick label style={font=\small},
    xmajorgrids=true,
    grid style={gray!20},
    title style={font=\bfseries},
    title={Overall Classification Accuracy by Model},
    legend style={at={(0.97,0.03)}, anchor=south east, font=\footnotesize},
    legend cell align={left},
]

\addplot[fill=darkblue!80, draw=darkblue] coordinates {
    (79.70,9) (75.80,8) (72.85,7)
};
\addlegendentry{Fine-tuned}

\addplot[fill=midblue!50, draw=midblue] coordinates {
    (65.85,6) (60.95,5) (56.45,4)
};
\addlegendentry{Commercial API}

\addplot[fill=lightblue!80, draw=lightblue!60!black] coordinates {
    (51.90,3) (31.25,2) (31.10,1)
};
\addlegendentry{Open-source (local)}

\node[font=\footnotesize, anchor=west] at (axis cs:80.5,9) {79.7\%};
\node[font=\footnotesize, anchor=west] at (axis cs:76.6,8) {75.8\%};
\node[font=\footnotesize, anchor=west] at (axis cs:73.7,7) {72.9\%};
\node[font=\footnotesize, anchor=west] at (axis cs:66.7,6) {65.9\%};
\node[font=\footnotesize, anchor=west] at (axis cs:61.8,5) {61.0\%};
\node[font=\footnotesize, anchor=west] at (axis cs:57.3,4) {56.5\%};
\node[font=\footnotesize, anchor=west] at (axis cs:52.7,3) {51.9\%};
\node[font=\footnotesize, anchor=west] at (axis cs:32.1,2) {31.3\%};
\node[font=\footnotesize, anchor=west] at (axis cs:31.9,1) {31.1\%};

\end{axis}
\end{tikzpicture}
\caption{Overall classification accuracy across nine models evaluated on a stratified 2,000-event sample from the GTD test set. The three fine-tuned models (dark bars) form a distinct performance tier above the commercial APIs (medium bars), which in turn outperform the locally deployed open-source models used without any fine-tuning (light bars). The gap between the weakest fine-tuned model (ConflLlama, 72.9\%) and the strongest commercial API (Gemini Flash, 65.9\%) is nearly seven percentage points.}
\label{fig:llm-accuracy}
\end{figure}

Figure~\ref{fig:llm-accuracy} makes the performance hierarchy visually concrete.

The gap between the three tiers is not merely statistically significant; it is substantively consequential. A model at 31\% accuracy misclassifies more than two-thirds of all events. Even at 66\%, the best commercial API introduces classification errors at roughly twice the rate of the fine-tuned models. For any research design that relies on automated event coding to construct variables for statistical analysis, error rates of this magnitude would introduce substantial measurement noise and potentially bias downstream estimates.

It is important to emphasize what this comparison does \textit{not} show. It does not demonstrate that commercial LLMs are inherently incapable of performing conflict event classification. With careful prompt engineering, few-shot examples, chain-of-thought reasoning, or retrieval-augmented generation, these models' performance would likely improve. But such optimization requires precisely the kind of iterative, task-specific investment that undermines the appeal of the ``buy'' option in the first place. If a researcher must spend weeks engineering prompts and validating outputs, the effort may approach or exceed what is required to simply fine-tune a model on labeled data, with the fine-tuned model offering superior performance and full reproducibility at the end of the process.

\subsubsection{Per-Class Performance}

The aggregate accuracy numbers conceal important variation across attack types. The per-class F1 scores reveal that zero-shot LLMs perform reasonably well on some categories and catastrophically poorly on others (the full per-class breakdown is reported in Table~\ref{tab:appendix-f1} and Figure~\ref{fig:appendix-heatmap} in the Appendix).

Two patterns are particularly noteworthy. First, on the most common category (Bombing/\allowbreak{}Explosion), Gemini 3 Flash achieves an F1 of 0.95, essentially matching the fine-tuned models (0.96 for ConfliBERT, 0.95 for Confli-mBERT). This confirms that for well-represented categories with distinctive linguistic markers (words like ``bomb,'' ``explosion,'' ``detonated'' are strong signals), even zero-shot classification can be effective. Second, on the ``Unknown'' category, \textit{all} zero-shot models collapse: F1 scores of 0.03 (Claude Haiku), 0.07 (Gemini Flash), and 0.09 (DeepSeek V3). This makes intuitive sense: ``Unknown'' is a residual category defined by the \textit{absence} of discriminating features, and a zero-shot model has no way to learn what human coders mean by ``Unknown'' without exposure to labeled examples. The fine-tuned models, by contrast, achieve F1 scores of 0.65 (ConfliBERT) and 0.61 (Confli-mBERT) on this category, having learned the implicit decision boundary from training data.

\subsubsection{Model Size versus Performance}

\begin{figure}[hbt!]
\centering
\begin{tikzpicture}
\begin{axis}[
    width=0.92\textwidth,
    height=9cm,
    xlabel={Model Size (Billion Parameters, log scale)},
    ylabel={Micro F1 Score},
    xmode=log,
    xmin=0.05, xmax=2500,
    ymin=0, ymax=1.1,
    grid=both,
    grid style={gray!20},
    title style={font=\bfseries},
    title={Model Size vs.\ Classification Performance},
    legend style={at={(0.97,0.03)}, anchor=south east, font=\footnotesize},
    legend cell align={left},
    xtick={0.1,1,10,100,1000},
    xticklabels={100M,1B,10B,100B,1T},
]

\addplot[only marks, mark=star, mark size=7pt, color=darkblue, thick] coordinates {
    (0.110, 0.831)
    (0.177, 0.797)
    (8.0, 0.759)
};
\addlegendentry{Fine-tuned}

\addplot[only marks, mark=diamond*, mark size=6pt, color=midblue!70!black] coordinates {
    (15.0, 0.664)
    (300.0, 0.730)
    (685.0, 0.665)
};
\addlegendentry{Commercial API}

\addplot[only marks, mark=*, mark size=5pt, color=lightblue!60!black] coordinates {
    (8.0, 0.348)
    (9.0, 0.334)
    (14.0, 0.555)
};
\addlegendentry{Open-source (local)}

\node[font=\footnotesize, anchor=south west, color=darkblue!80!black] at (axis cs:0.110, 0.845) {ConfliBERT};
\node[font=\footnotesize, anchor=north west, color=darkblue!80!black] at (axis cs:0.190, 0.785) {Confli-mBERT};
\node[font=\footnotesize, anchor=south, color=darkblue!80!black] at (axis cs:8.0, 0.775) {ConflLlama};

\node[font=\footnotesize, anchor=south west, color=midblue!70!black] at (axis cs:16, 0.674) {Claude Haiku};
\node[font=\footnotesize, anchor=south, color=midblue!70!black] at (axis cs:300, 0.745) {Gemini Flash};
\node[font=\footnotesize, anchor=east, color=midblue!70!black] at (axis cs:650, 0.650) {DeepSeek V3};

\node[font=\footnotesize, anchor=south, color=lightblue!60!black] at (axis cs:8.0, 0.360) {Llama3.1};
\node[font=\footnotesize, anchor=north, color=lightblue!60!black] at (axis cs:9.0, 0.320) {Gemma2};
\node[font=\footnotesize, anchor=south west, color=lightblue!60!black] at (axis cs:15, 0.565) {Qwen2.5};

\draw[->, thick, gray] (axis cs:1.5, 0.97) -- (axis cs:0.15, 0.86);
\node[font=\footnotesize, anchor=west, align=left, color=gray!70!black] at (axis cs:1.8, 0.98) {\textit{Small fine-tuned models outperform LLMs 100--6000$\times$ larger}};

\end{axis}
\end{tikzpicture}
\caption{Model size (total parameters, log scale) plotted against Micro F1 classification performance. Fine-tuned models ($\bigstar$) cluster in the upper left: small models with high performance. Commercial APIs ($\blacklozenge$) and open-source models ($\bullet$) occupy the lower right: large models with mediocre to poor performance. ConfliBERT, at 110 million parameters, outperforms DeepSeek V3.2, a model roughly 6,200 times its size. The relationship between size and performance is \textit{negative} across model families, demonstrating that task-specific fine-tuning overwhelmingly dominates model scale for classification tasks.}
\label{fig:size-perf}
\end{figure}
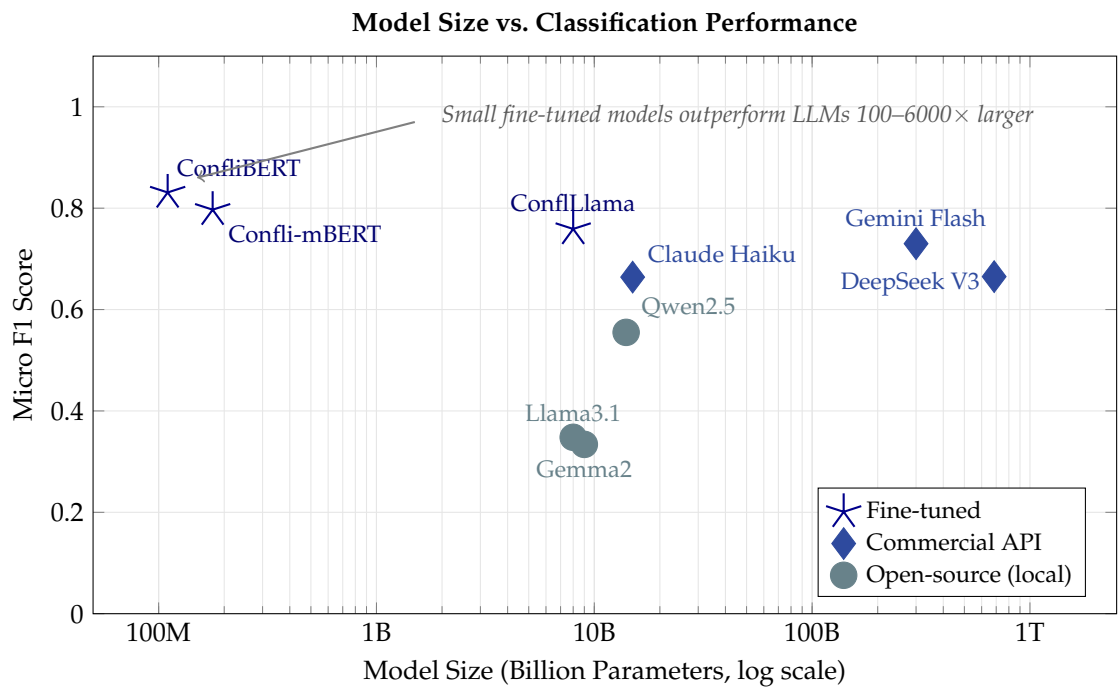

A natural hypothesis is that larger models should perform better at zero-shot classification, since they have been trained on more data and can represent more complex linguistic patterns. Figure~\ref{fig:size-perf} tests this hypothesis directly by plotting model size against classification performance.

The figure reveals a striking and counterintuitive pattern. The three best-performing models (ConfliBERT at 110M parameters, Confli-mBERT at 177M, and ConflLlama at 8B) are the three \textit{smallest} models in their respective architectural classes, clustered in the upper-left corner of the plot. The largest model in the evaluation, DeepSeek V3.2 at an estimated 685 billion parameters, achieves a Micro F1 of only 0.665, substantially below ConfliBERT's 0.831 despite being more than 6,000 times larger.

This result has a clear interpretation. For classification tasks with labeled training data, the relevant axis of variation is not model size but \textit{task-specific adaptation}. A 110-million-parameter model that has been trained on thousands of labeled examples of the target task will develop finely calibrated decision boundaries that a 685-billion-parameter model, operating from a general-purpose prompt, simply cannot replicate. The enormous parametric capacity of modern LLMs is an asset for tasks that require broad world knowledge, creative generation, or multi-step reasoning; for a focused classification task with well-defined categories, that capacity is mostly irrelevant.

The practical implication for political scientists is direct: purchasing access to a large commercial LLM is not a substitute for fine-tuning. A researcher with a laptop and a free Hugging Face account can produce a classifier that outperforms the most expensive API available, provided she has labeled training data. The barrier to entry is not compute or money; it is labeled data and the (modest) expertise required to run a fine-tuning script.

\subsubsection{The Cost of Buying}

Even setting aside the performance deficit, the cost structure of API-based classification presents a non-trivial barrier for the scale of analysis typical in political science. Table~\ref{tab:llm-cost} reports the per-token pricing and estimated costs for classifying the GTD at increasing scales.

\begin{table}[hbt!]
\centering
\caption{Commercial API Pricing and Projected Classification Costs (February 2026)}
\label{tab:llm-cost}
\begin{threeparttable}
\begin{tabular}{lrrrrr}
\toprule
\textbf{Model} & \textbf{Input} & \textbf{Output} & \textbf{2,000} & \textbf{37,709} & \textbf{170,623} \\
               & \textbf{(\$/M tok.)} & \textbf{(\$/M tok.)} & \textbf{rows} & \textbf{rows}$^a$ & \textbf{rows}$^b$ \\
\midrule
Claude Haiku 4.5 & \$1.00 & \$5.00 & \$1.00 & \$18.85 & \$85.31 \\
Gemini 3 Flash   & \$0.50 & \$3.00 & \$0.53 & \$9.99  & \$45.22 \\
DeepSeek V3.2    & \$0.26 & \$0.38 & \$0.20 & \$3.77  & \$17.06 \\
\midrule
\textbf{3-model total} & --- & --- & \textbf{\$1.73} & \textbf{\$32.61} & \textbf{\$147.59} \\
\bottomrule
\end{tabular}
\begin{tablenotes}
\small
\item $^a$ Full GTD test set (2017+). $^b$ Full GTD corpus (including training set).
\item Estimates assume $\sim$350 input tokens (system prompt + text) and $\sim$30 output tokens per request.
\item Costs are per \textit{single pass}; prompt iteration, ensemble methods, or chain-of-thought reasoning would multiply costs proportionally.
\end{tablenotes}
\end{threeparttable}
\end{table}

At first glance, the per-sample costs appear modest: classifying the full GTD test set with all three API models costs approximately \$33, and even the entire corpus costs under \$150. These figures, however, substantially understate the true cost of an API-based research workflow for three reasons.

First, research workflows require \textit{repeated} classification runs. The prompt used in this evaluation was itself the product of iterative refinement: testing different instruction formats, adjusting output parsing, and validating against known labels. In practice, a researcher developing an API-based classification pipeline will run the same data through the API dozens of times during development. Each iteration incurs the full cost again. A realistic estimate of the total API expenditure for a single project (including development, validation, and final production runs) would be 5--20$\times$ the single-pass cost shown in Table~\ref{tab:llm-cost}.

Second, the costs scale multiplicatively with the number of models evaluated. A responsible comparison, of the kind conducted here or recommended by best practices in computational social science \citep{grimmer2022text}, requires testing multiple providers and models rather than relying on a single API. Each additional model multiplies the total cost, creating a perverse incentive to evaluate fewer alternatives and potentially miss the best-performing option.

Third, and most critically for academic research, API pricing is \textit{non-stationary}. The prices in Table~\ref{tab:llm-cost} are snapshots from February 2026 via the OpenRouter aggregation service. Over the preceding 12 months, input token prices for the models tested had changed by factors of 1.5--3$\times$. There is no guarantee that a model available today will be available, or priced identically, when a reviewer asks for a replication, a co-author wants to extend the analysis, or a reader attempts to reproduce the results five years hence. By contrast, a fine-tuned model downloaded to a local machine remains available indefinitely at zero marginal cost per classification.

For context, fine-tuning Confli-mBERT, the model that outperforms every API tested, cost approximately \$5 to \$15 in cloud GPU time, or nothing at all on a university cluster. The fine-tuned model file is 700~MB, runs on consumer hardware, and can classify the entire GTD in under six minutes. The total cost of the fine-tuning approach, from start to finish, is less than the cost of a single API evaluation pass on the full corpus.

\subsubsection{Structural Concerns for Academic Research}

The performance and cost disadvantages documented above are compounded by structural concerns that go beyond dollars and accuracy points. These concerns are specific to the use of commercial APIs in the context of academic research, where reproducibility, data governance, and long-term availability are foundational requirements.

\begin{enumerate}[leftmargin=*]
    \item \textbf{Non-reproducibility.} Commercial LLMs are updated, retrained, and occasionally deprecated without notice. A classification pipeline that produces specific results with Claude Haiku 4.5 in February 2026 may produce different results when executed one month later, even with identical inputs and temperature set to zero. Model providers offer no guarantees of behavioral stability across versions, and versGiven their black-box nature, researchers cannot inspect their training data to determine if a given corpus was included in their pretraining, raising concerns about data contamination and reproducibility \citep{spirling2023world}. Third, there are significant privacy and security implications. of computational reproducibility that underpins quantitative social science.

    \item \textbf{Data sovereignty and ethics.} Submitting research data to external API endpoints raises questions that are particularly acute for conflict research. Event descriptions in the GTD contain sensitive information about attacks, victims, and perpetrators. Sending this data to servers operated by foreign commercial entities may conflict with institutional data governance policies, European GDPR requirements, or the terms of data use agreements with data providers. While most API providers offer data processing agreements and enterprise tiers with enhanced privacy guarantees, the complexity of navigating these requirements adds overhead that local inference on a university-owned machine avoids entirely.

    \item \textbf{Throughput and reliability.} API services impose rate limits that constrain throughput and introduce unpredictable delays. During the evaluation conducted for this paper, classifying 2,000 events across five API models required approximately 77 minutes of wall-clock time with 10 parallel workers. At this rate, processing the full GTD corpus ($\sim$170,000 events) with a single model would require roughly 14 hours of continuous operation, assuming no rate limit throttling, network interruptions, HTTP timeouts, or service degradation. In practice, the evaluation required implementing robust retry logic, checkpoint saving, and error handling to manage transient API failures. By contrast, Confli-mBERT classified all 37,709 test events locally in 5.5 minutes using Apple's Metal GPU acceleration on a standard consumer Macbook.

    \item \textbf{Pricing and availability risk.} The commercial LLM market is characterized by rapid consolidation, model deprecation, and price volatility. Models that are available and affordable today may be discontinued, repriced, or absorbed into different product tiers at any time. A research project that begins with one set of API costs may face substantially different costs by the time it reaches publication. Fine-tuning a local model involves a one-time fixed cost with no ongoing pricing exposure and no dependency on the continued existence of a commercial service.
\end{enumerate}

\subsubsection{Implications for the Build-Borrow-Buy Framework}

The evidence from this evaluation is unambiguous in its practical implications. For political science text classification tasks where labeled training data exists or can be created, commercial LLM APIs represent the least attractive of the three options in the build-borrow-buy framework. They are more expensive than local fine-tuning (\$33--\$150 per pass versus a one-time \$5--\$15), less accurate than any fine-tuned alternative (65.9\% for the best API vs.\ 72.9\% for the weakest fine-tuned model), and introduce dependencies on external services that undermine every pillar of responsible academic research: reproducibility, data governance, cost predictability, and long-term availability.

The one scenario in which API-based classification may be justified is as a rapid prototyping tool during the earliest stages of research, before labeled training data has been assembled for fine-tuning. A researcher exploring whether a text classification approach is feasible for a new research question might reasonably use an API for initial exploration. But even in this scenario, locally deployed open-source models (run through tools like Ollama or LMStudio) offer comparable convenience without the cost, sovereignty, or reproducibility concerns. And the performance of these local models, as documented in this evaluation, should set expectations appropriately: zero-shot classification, whether via API or local deployment, is a starting point, not an endpoint.

\section{Discussion}\label{sec:discussion}

\subsection{A Decision Framework for Model Selection}

The results of this study, though drawn from a single application domain, illuminate a pattern that generalizes broadly. The relative advantage of domain-specific pretraining over fine-tuning is not uniform; it is systematically mediated by class prevalence. This regularity suggests a practical decision framework for political scientists choosing among NLP approaches for any text classification task.

The framework rests on three considerations.

The first is \textbf{category prevalence}. As the empirical results demonstrate, the performance gap between a domain-specific model like ConfliBERT and a fine-tuned general model like Confli-mBERT is a direct function of how much labeled data is available for each category. For categories represented by thousands of training examples, fine-tuning alone provides sufficient signal, and domain-specific pretraining adds comparatively little. For categories with only tens or hundreds of examples, domain pretraining provides informative priors that compensate for the scarcity of supervised signal. Researchers should therefore begin by examining the prevalence distribution of the categories they need to classify. If the categories of primary interest are well-represented in the training data, fine-tuning is likely sufficient. If the key categories are rare, a domain-specific model warrants serious consideration.

The second consideration is \textbf{error tolerance}. Different research designs have different sensitivities to classification error. Aggregate analyses, such as tracking the annual frequency of bombings across countries, are relatively robust to random classification noise: errors that do not correlate with the variables of interest will attenuate estimated relationships toward zero rather than generate spurious findings \citep{grimmer2022text}. The measurement error introduced by Confli-mBERT at 79.66\% accuracy is comparable to, and likely smaller than, the inter-coder disagreement rates documented for human-coded event data \citep{mikhaylov2012coder, weidmann2016micro}. For studies where individual event-level coding must be highly accurate (for example, detailed case studies of specific incidents), any automated approach should be supplemented with manual verification.

The third consideration is \textbf{available resources}. The resource asymmetry between pretraining and fine-tuning is dramatic. Pretraining ConfliBERT required curating a domain-specific corpus of 33 million tokens, training a custom tokenizer, and running multi-GPU infrastructure for an extended period, at an estimated cost of hundreds to thousands of dollars and months of expert labor. Fine-tuning Confli-mBERT required downloading a publicly available base model, preparing labeled data, and running a standard training script for approximately four hours on a single GPU, at a cost of \$5 to \$15. The difference in required expertise is at least as consequential as the difference in cost: a graduate student can learn to fine-tune a transformer in a workshop, while pretraining from scratch requires expertise in distributed computing, data engineering, and tokenizer design that takes months to acquire.

Taken together, these three factors define a decision space. For a political scientist studying aggregate trends in common event types with limited computational resources, fine-tuning is the clear choice. For a researcher who needs to reliably detect rare event categories and has access to domain-specific tools like ConfliBERT, using the domain-specific model is warranted. For most researchers, who fall somewhere between these poles, fine-tuning offers the best trade-off between performance, cost, and risk.

\subsection{In Defense of Domain-Specific Models}

It is important to be clear about what this paper does and does not argue. It does not argue that ConfliBERT is unnecessary, nor that domain-specific pretraining was a misguided investment. On the contrary: ConfliBERT outperforms Confli-mBERT on every single attack type except two (Kidnapping and Unknown), and its advantages on rare event categories are substantial and practically meaningful.

ConfliBERT's value extends beyond raw classification performance. As a community resource, it provides a shared benchmark against which new approaches can be evaluated; it embeds domain knowledge that benefits researchers who use it without fully understanding its internals; and it has catalyzed further work, including ConflLlama \citep{meher2025conflllama}, that has expanded the toolkit available to conflict researchers. Domain-specific models play an important role in the intellectual infrastructure of their fields, and their development should be encouraged and supported.

The argument of this paper is rather that the \textit{decision} about whether to use a domain-specific model should be conscious and informed rather than reflexive. If ConfliBERT exists and is available for a given task, there is no reason not to use it. But if a researcher is working on a task for which no domain-specific model exists, or is working in a subdomain (such as protest event coding, legislative speech classification, or treaty text analysis) that ConfliBERT does not cover, the relevant alternative is not ``build your own ConfliBERT.'' The relevant alternative is ``fine-tune the best available general-purpose model on your labeled data.'' This paper shows that the resulting model will often be good enough for the task at hand.

\subsection{Beyond Conflict Studies}

Although the empirical analysis in this paper focuses on conflict event classification, the underlying dynamics generalize to any domain where political scientists are considering NLP tools. The key insight is structural: the marginal value of domain-specific pretraining depends on the gap between what a general-purpose model already knows about the domain and what it would need to know for the specific task.

For domains that are linguistically close to general web text (political news analysis, social media sentiment, legislative speech), the gap is likely small, and fine-tuning should perform well. For domains with highly specialized terminologies that are underrepresented in web corpora (historical diplomatic correspondence, legal statutes in non-English languages, handwritten archival materials), the gap may be larger, and domain pretraining may offer more substantial benefits.

The general principle is one of diminishing returns: there is a curve that relates investment in model specialization to downstream performance, and that curve flattens as the investment increases. The first few hours of fine-tuning on labeled task data produce large gains. The next step, domain-adaptive pretraining, produces smaller but still meaningful gains. Full pretraining from scratch on curated domain text produces smaller gains still, at dramatically higher cost. The optimal point on this curve depends on the researcher's specific needs, and the framework proposed in this paper helps identify where that point lies.

\subsection{The Rising Floor and the Narrowing Gap}

A broader implication of these findings concerns the trajectory of general-purpose language models. When ConfliBERT was developed, the best available general-purpose model (BERT) had been trained on approximately 3.3 billion tokens. Today, ModernBERT has been trained on 2 trillion tokens, a 600-fold increase, with substantial architectural improvements. The ``vocabulary gap'' between general and domain-specific corpora has narrowed dramatically as a result. A model trained on 2 trillion tokens of diverse web text has encountered large quantities of conflict-related news, academic analysis, and policy discussion simply as a byproduct of training on a comprehensive sample of human language.

This trend has a predictable consequence: as general-purpose models improve, the marginal value of domain-specific pretraining shrinks. The next generation of encoder-only transformers will be trained on even larger corpora with further architectural advances, raising the fine-tuning baseline still higher. For the discipline, this means that the investment calculus will continue to shift in favor of fine-tuning over time. Rather than building new domain-specific models every few years to keep pace with architectural evolution, political scientists can adopt a strategy of fine-tuning the latest available general-purpose model on their task data, capturing the benefits of broader progress in the field without requiring specialized infrastructure.

This does not mean that domain-specific pretraining will become irrelevant. There will always be tasks and domains for which general models are insufficient. But the set of tasks for which fine-tuning is ``good enough'' will continue to grow, and the discipline should adjust its methodological guidance accordingly.

\subsection{Limitations}

Several limitations of this study should be acknowledged. First, the comparison is restricted to a single task (attack type classification on the GTD). ConfliBERT's advantages may be more pronounced for tasks that require deeper engagement with conflict-specific semantics, such as actor identification, event relevance filtering, or causal attribution. Future work should extend this comparison to a broader suite of NLP tasks across multiple subfields of political science.

Second, I have not exhaustively optimized the fine-tuning procedure. Techniques such as class-weighted loss functions, threshold optimization per class, data augmentation for rare event types, or domain-adaptive continued pretraining of ModernBERT on conflict text could potentially narrow the gap with ConfliBERT. The results reported here therefore represent a \textit{lower bound} on what fine-tuning can achieve.

Third, the temporal split (pre-2017 vs.\ 2017+) means that distributional shifts in attack patterns across time may affect both models differently in ways not fully captured by the evaluation. If certain attack modalities have become more or less common after 2017, performance comparisons may partly reflect these shifts rather than pure model quality.

Fourth, I have relied on published results for ConfliBERT's performance rather than retraining it from scratch under identical conditions. Differences in preprocessing, tokenization details, or evaluation methodology between the original ConfliBERT study and the present work could contribute to some portion of the observed performance gap.

\section{Conclusion}\label{sec:conclusion}

This paper began with a practical question that an increasing number of political scientists face: when adopting NLP tools for text classification (or any other similar task related to textual data), how much should you invest in model specialization? The answer, grounded in a systematic comparison of a domain-specific pretrained model (ConfliBERT) and a fine-tuned general-purpose model (Confli-mBERT) on conflict event classification, is that the right level of investment depends on the specific research question being asked.

For the common event types that dominate most empirical analyses, fine-tuning a modern general-purpose transformer produces results that are competitive with the gold standard at a fraction of the cost and complexity. On bombings, armed assaults, kidnappings, and facility attacks, the two models are nearly indistinguishable. For rare event categories, ConfliBERT's domain-specific pretraining confers clear advantages, demonstrating that specialized models remain valuable tools for specific research needs.

The decision framework developed in this paper reduces the model selection problem to three questions. What is the prevalence distribution of the categories you need to classify? How sensitive is your downstream analysis to classification error? And what computational and human resources are available? For most political scientists working with common event types and conducting aggregate analyses, fine-tuning will provide the best balance of performance, cost, and reproducibility. For researchers studying rare events or requiring event-level precision, domain-specific models like ConfliBERT are the appropriate tool.

The broader message extends well beyond conflict studies. As AI tools become standard equipment across political science, the discipline needs a mature conversation about how to choose among them. That conversation should be grounded not in which tool is ``best'' in the abstract, but in which tool is best \textit{for the task at hand}. The evidence presented here suggests that the answer, more often than political scientists might expect, is the simplest and most accessible option available.

As general-purpose language models continue to improve, the floor from which fine-tuning begins will continue to rise. The discipline should prepare for a future in which the practical recommendation is increasingly straightforward: start with fine-tuning, validate against your specific requirements, and invest in specialization only when the evidence justifies it. For the majority of applications, the data matters more than the model.

\paragraph{Acknowledgments}
I am grateful for the technical assistance of the Event Data group across the University of Texas at Dallas, University of Arizona, West Virginia University and King Saud University.

\paragraph{Funding Statement}
This research was supported by NSF Award 2311142 and utilized Delta resources at NCSA/University of Illinois (Allocation CIS220162) through the Advanced Cyberinfrastructure Coordination Ecosystem (ACCESS) program.

\paragraph{Competing Interests}
No competing interests.

\paragraph{Data Availability Statement}
The Confli-mBERT model, training code, and evaluation data are publicly available at \url{https://huggingface.co/shreyasmeher/Confli-mBERT} along with a forthcoming Dataverse repo.

\newpage
\bibliographystyle{unsrtnat}
\bibliography{references}

\newpage
\appendix

\section{Extended LLM Evaluation Results}\label{app:llm}

\begin{figure}[hbt!]
\centering
\begin{tikzpicture}
\begin{axis}[
    width=0.75\textwidth,
    height=8.5cm,
    colormap={GreenRed}{
        rgb255(0cm)=(215,48,39)
        rgb255(1cm)=(244,109,67)
        rgb255(2cm)=(253,174,97)
        rgb255(3cm)=(254,224,139)
        rgb255(4cm)=(217,239,139)
        rgb255(5cm)=(166,217,106)
        rgb255(6cm)=(102,189,99)
        rgb255(7cm)=(26,152,80)
    },
    colorbar,
    colorbar style={
        ylabel={F1 Score},
    },
    point meta min=0,
    point meta max=1,
    xtick={0,1,2,3,4,5,6,7,8},
    xticklabels={
        {ConfliBERT},
        {Confli-mBERT},
        {ConflLlama},
        {Gemma2},
        {Qwen2.5},
        {Llama3.1},
        {Claude Haiku 4.5},
        {Gemini 3 Flash},
        {DeepSeek V3.2}
    },
    x tick label style={rotate=45, anchor=east, font=\footnotesize},
    ytick={0,1,2,3,4,5,6,7,8},
    yticklabels={
        {Assassination},
        {Armed Assault},
        {Bombing/Explosion},
        {Hijacking},
        {Barricade Incident},
        {Kidnapping},
        {Facility Attack},
        {Unarmed Assault},
        {Unknown}
    },
    y tick label style={font=\footnotesize},
    title style={font=\bfseries},
    title={Per-Class F1 Scores Across All Evaluated Models},
    xlabel={},
    ylabel={},
    enlargelimits=false,
    axis on top,
]

\addplot[
    matrix plot*,
    mesh/cols=9,
    mesh/rows=9,
    point meta=explicit,
] table[meta=C] {
x y C
0 0 0.79
1 0 0.64
2 0 0.66
3 0 0.18
4 0 0.32
5 0 0.19
6 0 0.52
7 0 0.59
8 0 0.49
0 1 0.75
1 1 0.71
2 1 0.70
3 1 0.41
4 1 0.56
5 1 0.37
6 1 0.60
7 1 0.63
8 1 0.57
0 2 0.96
1 2 0.95
2 2 0.91
3 2 0.48
4 2 0.80
5 2 0.55
6 2 0.87
7 2 0.95
8 2 0.91
0 3 0.67
1 3 0.43
2 3 0.50
3 3 0.13
4 3 0.30
5 3 0.14
6 3 0.55
7 3 0.77
8 3 0.50
0 4 0.35
1 4 0.00
2 4 0.25
3 4 0.11
4 4 0.11
5 4 0.03
6 4 0.35
7 4 0.38
8 4 0.24
0 5 0.91
1 5 0.93
2 5 0.79
3 5 0.31
4 5 0.67
5 5 0.32
6 5 0.82
7 5 0.85
8 5 0.85
0 6 0.74
1 6 0.78
2 6 0.70
3 6 0.11
4 6 0.33
5 6 0.17
6 6 0.65
7 6 0.70
8 6 0.62
0 7 0.55
1 7 0.19
2 7 0.48
3 7 0.04
4 7 0.13
5 7 0.02
6 7 0.34
7 7 0.50
8 7 0.18
0 8 0.65
1 8 0.61
2 8 0.49
3 8 0.17
4 8 0.12
5 8 0.13
6 8 0.03
7 8 0.07
8 8 0.09
};

\node[font=\tiny, white] at (axis cs:0,0) {.79};
\node[font=\tiny, white] at (axis cs:1,0) {.64};
\node[font=\tiny, white] at (axis cs:2,0) {.66};
\node[font=\tiny, white] at (axis cs:3,0) {.18};
\node[font=\tiny, white] at (axis cs:4,0) {.32};
\node[font=\tiny, white] at (axis cs:5,0) {.19};
\node[font=\tiny, white] at (axis cs:6,0) {.52};
\node[font=\tiny, white] at (axis cs:7,0) {.59};
\node[font=\tiny, white] at (axis cs:8,0) {.49};
\node[font=\tiny, white] at (axis cs:0,1) {.75};
\node[font=\tiny, white] at (axis cs:1,1) {.71};
\node[font=\tiny, white] at (axis cs:2,1) {.70};
\node[font=\tiny] at (axis cs:3,1) {.41};
\node[font=\tiny] at (axis cs:4,1) {.56};
\node[font=\tiny] at (axis cs:5,1) {.37};
\node[font=\tiny, white] at (axis cs:6,1) {.60};
\node[font=\tiny, white] at (axis cs:7,1) {.63};
\node[font=\tiny, white] at (axis cs:8,1) {.57};
\node[font=\tiny, white] at (axis cs:0,2) {.96};
\node[font=\tiny, white] at (axis cs:1,2) {.95};
\node[font=\tiny, white] at (axis cs:2,2) {.91};
\node[font=\tiny] at (axis cs:3,2) {.48};
\node[font=\tiny, white] at (axis cs:4,2) {.80};
\node[font=\tiny, white] at (axis cs:5,2) {.55};
\node[font=\tiny, white] at (axis cs:6,2) {.87};
\node[font=\tiny, white] at (axis cs:7,2) {.95};
\node[font=\tiny, white] at (axis cs:8,2) {.91};
\node[font=\tiny, white] at (axis cs:0,3) {.67};
\node[font=\tiny] at (axis cs:1,3) {.43};
\node[font=\tiny] at (axis cs:2,3) {.50};
\node[font=\tiny, white] at (axis cs:3,3) {.13};
\node[font=\tiny] at (axis cs:4,3) {.30};
\node[font=\tiny, white] at (axis cs:5,3) {.14};
\node[font=\tiny, white] at (axis cs:6,3) {.55};
\node[font=\tiny, white] at (axis cs:7,3) {.77};
\node[font=\tiny] at (axis cs:8,3) {.50};
\node[font=\tiny] at (axis cs:0,4) {.35};
\node[font=\tiny, white] at (axis cs:1,4) {.00};
\node[font=\tiny] at (axis cs:2,4) {.25};
\node[font=\tiny, white] at (axis cs:3,4) {.11};
\node[font=\tiny, white] at (axis cs:4,4) {.11};
\node[font=\tiny, white] at (axis cs:5,4) {.03};
\node[font=\tiny] at (axis cs:6,4) {.35};
\node[font=\tiny] at (axis cs:7,4) {.38};
\node[font=\tiny] at (axis cs:8,4) {.24};
\node[font=\tiny, white] at (axis cs:0,5) {.91};
\node[font=\tiny, white] at (axis cs:1,5) {.93};
\node[font=\tiny, white] at (axis cs:2,5) {.79};
\node[font=\tiny] at (axis cs:3,5) {.31};
\node[font=\tiny, white] at (axis cs:4,5) {.67};
\node[font=\tiny] at (axis cs:5,5) {.32};
\node[font=\tiny, white] at (axis cs:6,5) {.82};
\node[font=\tiny, white] at (axis cs:7,5) {.85};
\node[font=\tiny, white] at (axis cs:8,5) {.85};
\node[font=\tiny, white] at (axis cs:0,6) {.74};
\node[font=\tiny, white] at (axis cs:1,6) {.78};
\node[font=\tiny, white] at (axis cs:2,6) {.70};
\node[font=\tiny, white] at (axis cs:3,6) {.11};
\node[font=\tiny] at (axis cs:4,6) {.33};
\node[font=\tiny, white] at (axis cs:5,6) {.17};
\node[font=\tiny, white] at (axis cs:6,6) {.65};
\node[font=\tiny, white] at (axis cs:7,6) {.70};
\node[font=\tiny, white] at (axis cs:8,6) {.62};
\node[font=\tiny, white] at (axis cs:0,7) {.55};
\node[font=\tiny, white] at (axis cs:1,7) {.19};
\node[font=\tiny] at (axis cs:2,7) {.48};
\node[font=\tiny, white] at (axis cs:3,7) {.04};
\node[font=\tiny, white] at (axis cs:4,7) {.13};
\node[font=\tiny, white] at (axis cs:5,7) {.02};
\node[font=\tiny] at (axis cs:6,7) {.34};
\node[font=\tiny] at (axis cs:7,7) {.50};
\node[font=\tiny, white] at (axis cs:8,7) {.18};
\node[font=\tiny, white] at (axis cs:0,8) {.65};
\node[font=\tiny, white] at (axis cs:1,8) {.61};
\node[font=\tiny] at (axis cs:2,8) {.49};
\node[font=\tiny, white] at (axis cs:3,8) {.17};
\node[font=\tiny, white] at (axis cs:4,8) {.12};
\node[font=\tiny, white] at (axis cs:5,8) {.13};
\node[font=\tiny, white] at (axis cs:6,8) {.03};
\node[font=\tiny, white] at (axis cs:7,8) {.07};
\node[font=\tiny, white] at (axis cs:8,8) {.09};

\end{axis}
\end{tikzpicture}
\caption{F1 scores for all nine models across all nine attack types. Green cells indicate strong performance ($\geq$0.70); yellow indicates moderate performance (0.40--0.69); red indicates poor performance ($<$0.40). The leftmost three columns (fine-tuned models) are predominantly green across most attack types. The rightmost six columns (zero-shot models) show high variance: strong performance on Bombing/Explosion but near-zero performance on Unknown and Barricade Incident.}
\label{fig:appendix-heatmap}
\end{figure}
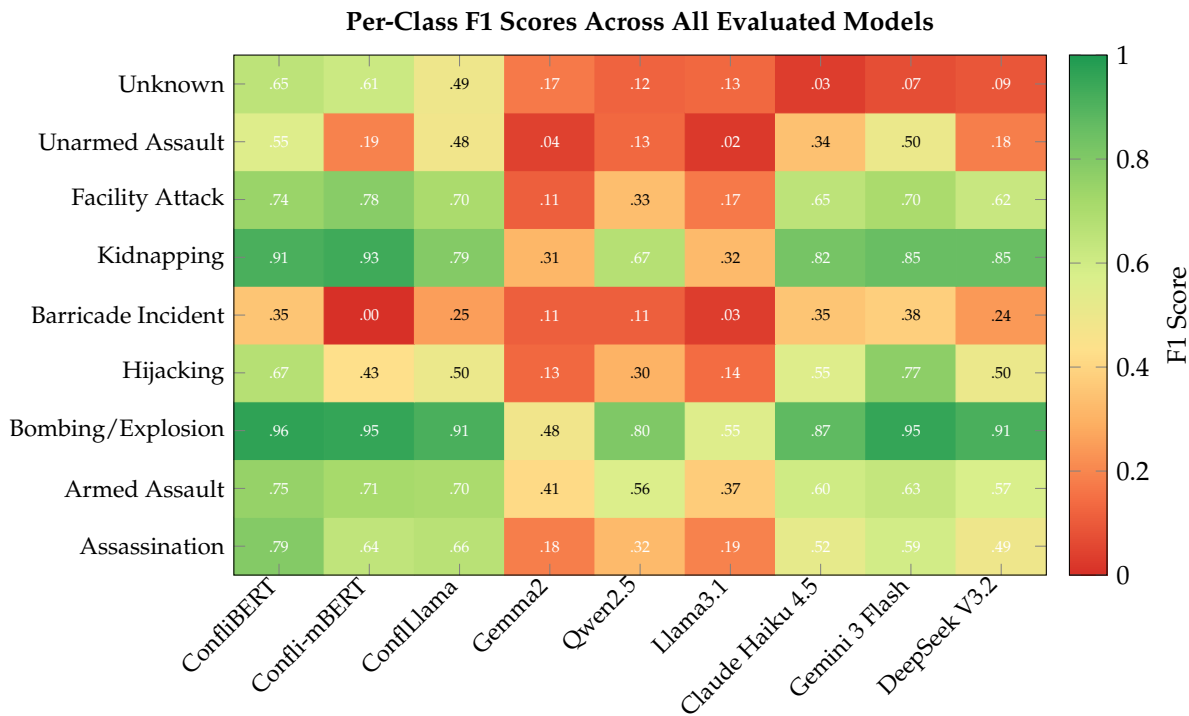

This appendix presents the full per-class breakdown of the nine-model evaluation described in Section~5.6. Figure~\ref{fig:appendix-heatmap} visualizes the F1 scores as a color-coded heatmap, and Table~\ref{tab:appendix-f1} reports the exact values.

Several patterns are visible in the heatmap. First, the Bombing/Explosion row is nearly uniformly green across all models, confirming that this category's distinctive linguistic markers make it classifiable even without task-specific training. Gemini Flash (0.95) and DeepSeek V3 (0.91) match or approach the fine-tuned models on this category alone. Second, the Unknown row is nearly uniformly red for zero-shot models, with F1 scores below 0.13 for all six. This residual category is effectively invisible to models that have not been trained to recognize it. Third, the Kidnapping row reveals an interesting pattern: the commercial APIs (0.82--0.85) substantially outperform the local open-source models (0.31--0.67), suggesting that the commercial models' broader training data includes sufficient exposure to kidnapping-related language to support reasonable classification even without fine-tuning.

Table~\ref{tab:appendix-f1} reports the exact F1 scores underlying the heatmap.

\begin{table}[hbt!]
\centering
\caption{Per-Class F1 Scores by Attack Type (All Nine Models)}
\label{tab:appendix-f1}
\resizebox{\textwidth}{!}{
\begin{tabular}{lrrr|rrr|rrr}
\toprule
& \multicolumn{3}{c|}{\textbf{Fine-tuned}} & \multicolumn{3}{c|}{\textbf{Commercial API}} & \multicolumn{3}{c}{\textbf{Open-source (local)}} \\
\textbf{Attack Type} & \textbf{ConfliBERT} & \textbf{Confli-mBERT} & \textbf{ConflLlama} & \textbf{Gemini Flash} & \textbf{Claude Haiku} & \textbf{DeepSeek V3} & \textbf{Qwen2.5} & \textbf{Llama3.1} & \textbf{Gemma2} \\
\midrule
Bombing/Explosion  & \textbf{0.96} & 0.95 & 0.91 & 0.95 & 0.87 & 0.91 & 0.80 & 0.55 & 0.48 \\
Kidnapping         & 0.91 & \textbf{0.93} & 0.79 & 0.85 & 0.82 & 0.85 & 0.67 & 0.32 & 0.31 \\
Assassination      & \textbf{0.79} & 0.64 & 0.66 & 0.59 & 0.52 & 0.49 & 0.32 & 0.19 & 0.18 \\
Armed Assault      & \textbf{0.75} & 0.71 & 0.70 & 0.63 & 0.60 & 0.57 & 0.56 & 0.37 & 0.41 \\
Facility Attack    & 0.74 & \textbf{0.78} & 0.70 & 0.70 & 0.65 & 0.62 & 0.33 & 0.17 & 0.11 \\
Hijacking          & 0.67 & 0.43 & 0.50 & \textbf{0.77} & 0.55 & 0.50 & 0.30 & 0.14 & 0.13 \\
Unknown            & \textbf{0.65} & 0.61 & 0.49 & 0.07 & 0.03 & 0.09 & 0.12 & 0.13 & 0.17 \\
Unarmed Assault    & \textbf{0.55} & 0.19 & 0.48 & 0.50 & 0.34 & 0.18 & 0.13 & 0.02 & 0.04 \\
Barricade Incident & 0.35 & 0.00 & 0.25 & \textbf{0.38} & 0.35 & 0.24 & 0.11 & 0.03 & 0.11 \\
\midrule
\textbf{Macro Avg} & \textbf{0.708} & 0.580 & 0.610 & 0.602 & 0.526 & 0.492 & 0.371 & 0.214 & 0.215 \\
\bottomrule
\end{tabular}
}
\end{table}

A notable finding in Table~\ref{tab:appendix-f1} is that Gemini Flash achieves the \textit{highest} F1 score of any model on Hijacking (0.77) and Barricade Incident (0.38), categories where even ConfliBERT achieves only 0.67 and 0.35, respectively. This suggests that for very rare categories where fine-tuning data is extremely scarce (8 and 12 instances in the 2,000-row sample), the broader world knowledge of a large commercial LLM can occasionally compensate for the absence of task-specific training. However, these isolated advantages do not offset the systematic underperformance across all other categories, particularly the catastrophic failure on the Unknown category (0.07 vs.\ 0.65 for ConfliBERT).

\subsection{Classification Prompt}\label{app:prompt}

The prompt used for all six zero-shot LLM evaluations is reproduced below. Each model received an identical system instruction, five few-shot examples, and the target text. The API was called with \texttt{temperature=0} and \texttt{max\_tokens=150} to minimize output variability.

\begin{tcolorbox}[promptbox, title=System Prompt]
\small
You are an expert conflict event classifier trained on the Global Terrorism Database (GTD). Your task is to classify incident descriptions into attack type categories.

\medskip
You MUST classify each incident into exactly the categories from this list:

\begin{enumerate}[leftmargin=2em, itemsep=0pt, topsep=2pt]
  \item Assassination
  \item Armed Assault
  \item Bombing/Explosion
  \item Hijacking
  \item Hostage Taking (Barricade Incident)
  \item Hostage Taking (Kidnapping)
  \item Facility/Infrastructure Attack
  \item Unarmed Assault
  \item Unknown
\end{enumerate}

\medskip
\textbf{Rules:}
\begin{itemize}[leftmargin=2em, itemsep=0pt, topsep=2pt]
  \item Output ONLY a valid JSON object mapping category names to probability scores.
  \item Each probability must be between 0.0 and 1.0; probabilities must sum to 1.0.
  \item Use ONLY category names from the list above (exact spelling and capitalization).
  \item An incident may involve multiple attack types. Assign probability mass accordingly.
  \item If the description is vague or unclear, use the ``Unknown'' category.
  \item Do NOT include any explanation, only the JSON object.
\end{itemize}
\end{tcolorbox}

\begin{tcolorbox}[promptbox, title=Few-Shot Examples (5 of 5 shown)]
\small
\textbf{Example 1} (single-label):\\[2pt]
\textit{``01/15/2018: Assailants detonated a vehicle-borne IED near a government checkpoint in Kabul, Afghanistan. At least 95 people were killed and 158 wounded. The Taliban claimed responsibility.''}\\[2pt]
$\rightarrow$ \texttt{\{"Bombing/Explosion": 1.0\}}

\medskip
\textbf{Example 2} (multi-label):\\[2pt]
\textit{``03/22/2017: Gunmen opened fire on unarmed civilians at a marketplace in Maiduguri, Nigeria, killing twelve. The attackers fled on motorcycles. Boko Haram was suspected.''}\\[2pt]
$\rightarrow$ \texttt{\{"Assassination": 0.4, "Armed Assault": 0.6\}}

\medskip
\textbf{Example 3} (kidnapping):\\[2pt]
\textit{``06/10/2019: Assailants abducted four foreign aid workers from their vehicle near the border region of Mali. The hostages were held for ransom.''}\\[2pt]
$\rightarrow$ \texttt{\{"Hostage Taking (Kidnapping)": 1.0\}}

\medskip
\textbf{Example 4} (infrastructure):\\[2pt]
\textit{``09/05/2017: Unknown assailants set fire to a telecommunications tower and destroyed electrical transformers in rural Colombia. The ELN was suspected.''}\\[2pt]
$\rightarrow$ \texttt{\{"Facility/Infrastructure Attack": 1.0\}}

\medskip
\textbf{Example 5} (complex multi-label):\\[2pt]
\textit{``11/28/2018: Armed militants stormed a hotel in Nairobi, Kenya, taking hostages and detonating explosives in the lobby. Security forces responded and a prolonged siege ensued. At least 21 killed. Al-Shabaab claimed responsibility.''}\\[2pt]
$\rightarrow$ \texttt{\{"Bombing/Explosion": 0.3,} \texttt{"Armed Assault": 0.3,} \texttt{"Hostage Taking (Barricade Incident)": 0.4\}}
\end{tcolorbox}

\begin{tcolorbox}[promptbox, title=API Configuration]
\small
All models were queried via the OpenRouter API (\texttt{openrouter.ai/api/v1/chat/completions}) with the following parameters: \texttt{temperature=0} (deterministic output), \texttt{max\_tokens=150}, \texttt{top\_p=1.0}. The message sequence consisted of the system prompt, followed by the five few-shot examples as alternating user/assistant turns, and finally the target text prefixed with ``Classify this incident:''.
\end{tcolorbox}

\end{document}